\documentclass[11pt,a4paper]{article}
\usepackage{times,latexsym}
\usepackage{url}
\usepackage[T1]{fontenc}
\usepackage[acceptedWithA]{tacl2021v1}

\usepackage[]{tacl2021v1}

\usepackage{xspace,mfirstuc,tabulary}

\usepackage{arydshln}
\usepackage{times}
\usepackage{latexsym}
\usepackage{adjustbox}
\usepackage{caption}
\usepackage{subcaption}
\usepackage{booktabs}
\usepackage{graphicx}
\usepackage{array}
\usepackage{multirow}
\usepackage{multicol}
\usepackage{booktabs}
\usepackage{pgfplots}
\usepackage{url}
\usepackage{colortbl}
\usepackage{enumitem}

\usepackage{amsmath}
\usepackage{amssymb}
\usepackage{algpseudocode}
\usepackage{algorithm}

\usepackage{pifont}
\usepackage[T1]{pbsi}
\newcommand{\cmark}{\ding{51}}%
\newcommand{\xmark}{\ding{55}}%
\makeatletter
\NewDocumentCommand{\brushfrac}{mm}{%
  \mbox{%
    \bsifamily
    \check@mathfonts
    \sbox\z@{/}%
    \raisebox{\dimeval{\ht\z@-\height}}{\fontsize{\sf@size}{0}\selectfont#1}%
    \kern-0.2em/\kern-0.2em
    \raisebox{-\dp\z@}{\fontsize{\sf@size}{0}\selectfont#2}%
  }%
}
\makeatother

\algnewcommand\algorithmicforeach{\textbf{for each}}
\algdef{S}[FOR]{ForEach}[1]{\algorithmicforeach\ #1\ \algorithmicdo}

\usepgfplotslibrary{groupplots}
\usetikzlibrary{pgfplots.groupplots}
\usetikzlibrary{plotmarks}
\usetikzlibrary{calc}
\usepgfplotslibrary{external}
\usetikzlibrary{positioning}
\usepackage[T1]{fontenc}
\usepackage{color,xcolor}
\usepackage{soul} 
\soulregister{\cite}7
\soulregister{\citep}7
\soulregister{\ref}7

\pgfdeclarelayer{foreground}
\pgfsetlayers{main,foreground}

\usetikzlibrary{patterns}
\pgfplotsset{compat=1.16}

\definecolor{lightblue}{RGB}{178,178,255}
\definecolor{darkgreen}{RGB}{51,153,51}
\definecolor{overlapcl}{RGB}{153,184,194}
\definecolor{forestgreen}{rgb}{0.13, 0.55, 0.13}
\colorlet{lightgrey}{white!90!black}
\colorlet{mixedcolor}{darkgreen!50!lightblue}

\newif\iftaclinstructions
\taclinstructionsfalse 
\iftaclinstructions
\renewcommand{\confidential}{}
\renewcommand{\anonsubtext}{(No author info supplied here, for consistency with
TACL-submission anonymization requirements)}
\newcommand{\instr}
\fi

\iftaclpubformat 

\else

\fi

\newcolumntype{C}[1]{>{\centering\arraybackslash}m{#1}}
\pgfplotsset{compat=1.18}

\newcommand{\DrawThreePercentageBar}[3]{%
\begin{tikzpicture}[
squarednode1/.style={rectangle, inner sep=0pt, draw=white!50, fill=black, thick, minimum width=0.8cm, minimum height = (#1/100.0)*2.0cm},
squarednode2/.style={rectangle, inner sep=0pt, draw=white!50, fill=red, thick, minimum width=0.8cm, minimum height = (#2/100.0)*2.0cm},
squarednode3/.style={rectangle, inner sep=0pt, draw=white!50, fill=blue, thick, minimum width=0.8cm, minimum height = (#3/100.0)*2.0cm},
]
\coordinate (A-Coord) at (0,0);
\coordinate (B-Coord) at (0.9,0);
\coordinate (C-Coord) at (1.8,0);
\node[squarednode1, anchor=south] (A-Node) at (A-Coord) {};
\node[squarednode2, anchor=south] (B-Node) at (B-Coord) {};
\node[squarednode3, anchor=south] (C-Node) at (C-Coord) {};
\node[anchor=north] at (A-Coord) {#1};
\node[anchor=north] at (B-Coord) {#2};
\node[anchor=north] at (C-Coord) {#3};
\end{tikzpicture}%
}

\newcommand{\cometDrawThreePercentageBar}[3]{%
\begin{tikzpicture}[
squarednode1/.style={rectangle, inner sep=0pt, draw=white!50, fill=black, thick, minimum width=0.8cm, minimum height = (#1/100.0)*1.0cm},
squarednode2/.style={rectangle, inner sep=0pt, draw=white!50, fill=red, thick, minimum width=0.8cm, minimum height = (#2/100.0)*1.0cm},
squarednode3/.style={rectangle, inner sep=0pt, draw=white!50, fill=blue, thick, minimum width=0.8cm, minimum height = (#3/100.0)*1.0cm},
]
\coordinate (A-Coord) at (0,0);
\coordinate (B-Coord) at (0.9,0);
\coordinate (C-Coord) at (1.8,0);
\node[squarednode1, anchor=south] (A-Node) at (A-Coord) {};
\node[squarednode2, anchor=south] (B-Node) at (B-Coord) {};
\node[squarednode3, anchor=south] (C-Node) at (C-Coord) {};
\node[anchor=north] at (A-Coord) {#1};
\node[anchor=north] at (B-Coord) {#2};
\node[anchor=north] at (C-Coord) {#3};
\end{tikzpicture}%
}

\title{Salute the Classic: Revisiting Challenges of Machine Translation \\ in the Age of Large Language Models}

\author{Jianhui Pang$^{1}$\thanks{~~Work was done when Jianhui Pang and Fanghua Ye were interning at Tencent AI Lab.}~~~~~Fanghua Ye$^{2}$\footnotemark[1]~~~~~\bf Derek Fai Wong$^{1}$\footnotemark[2]~~~~~\bf Dian Yu$^{3}$\\ \bf Shuming Shi$^{3}$~~~~~\bf Zhaopeng Tu$^{3}$~~~~~\bf Longyue Wang$^{3}$\thanks{~~Corresponding Authors.}\\$^1$University of Macau~~~~~~$^2$University College London~~~~~$^3$Tencent AI Lab\\\normalsize nlp2ct.pangjh3@gmail.com, fanghua.ye.19@ucl.ac.uk, derekfw@um.edu.mo \\ \normalsize \{yudian, shumingshi, zptu, vinnylywang\}@tencent.com}

\date{}

\begin{document}
\maketitle
\begin{abstract}


The evolution of Neural Machine Translation (NMT) has been significantly influenced by six core challenges \cite{koehn-knowles-2017-six}, which have acted as benchmarks for progress in this field. 
This study revisits these challenges, offering insights into their ongoing relevance in the context of advanced Large Language Models (LLMs):
\textit{domain mismatch}, \textit{amount of parallel data}, \textit{rare word prediction}, \textit{translation of long sentences}, \textit{attention model as word alignment}, and \textit{sub-optimal beam search}. 
Our empirical findings show that LLMs effectively reduce reliance on parallel data for major languages during pretraining and significantly improve translation of long sentences containing approximately 80 words, even translating documents up to 512 words. 
Despite these improvements, challenges in domain mismatch and rare word prediction persist. 
While NMT-specific challenges like word alignment and beam search may not apply to LLMs, we identify three new challenges in LLM-based translation: inference efficiency, translation of low-resource languages during pretraining, and human-aligned evaluation.

\end{abstract}

\section{Introduction}

\begin{table*}[t]
    \centering
    \resizebox{1.0\textwidth}{!}{
    \begin{tabular}{lp{3.2cm}ccp{7.5cm}}
    \toprule
     \multirow{2}{*}{\bf Challenges}  & \multirow{2}{*}{\bf Experiments} &  \multicolumn{3}{l}{\bf Outcomes}  \\ \cmidrule{3-5}
       &  & \bf Enc2Dec  & \bf LLMs & \bf Takeaways  \\ \midrule
     
     \bf Domain  & Llama2-7B/13B & \multirow{2}{*}{\xmark}   & \multirow{2}{*}{\xmark} & \multirow{2}{*}{\parbox{7.5cm}{LLMs demonstrate overall advancements yet still face significant domain variance issues.}}    \\ 

     \bf Mismatch & OPUS datasets & & &  \\ \cdashline{1-5}\noalign{\vskip 0.5ex}

     \bf Amount of   & Llama2-7B/13B & \multirow{2}{*}{\xmark}  & \multirow{2}{*}{\cmark} & \multirow{2}{*}{\parbox{7.5cm}{LLMs diminish the dependence on bilingual data for high-resource pretraining languages.}} \\

     \bf Parallel Data & WMT23 datasets & & &  \\ \cdashline{1-5}\noalign{\vskip 0.5ex}

     \bf Rare Word    & Llama2-7B/13B &  \multirow{2}{*}{\xmark}   & \multirow{2}{*}{\xmark} & \multirow{2}{*}{\parbox{7.5cm}{LLMs consistently struggle with predicting infrequent words.}}  \\ 

    \bf  Prediction & WMT23 datasets & & &  \\ \cdashline{1-5}\noalign{\vskip 0.5ex}
        
     \bf Long Sentence   & Llama2-7B/13B &  \multirow{3}{*}{\xmark}    & \multirow{3}{*}{\cmark} & \multirow{3}{*}{\parbox{7.5cm}{LLMs address the task of translating long sentences and display remarkable performance at the document level.}} \\ 

     \bf Translation & WMT23 datasets & & &  \\ 
     & & & & \\ \cdashline{1-5}\noalign{\vskip 0.5ex}

     \bf Word Alignment & Llama2-7B & \multirow{3}{*}{\cmark}  & \multirow{3}{*}{\xmark} & \multirow{3}{*}{\parbox{7.5cm}{The attention weights in LLMs are unsuitable for word alignment extraction, yet they provide valuable insights into model interpretability.}}  \\ 

     & WMT23 testsuits & & &  \\ 
     & & & & \\ \cdashline{1-5}\noalign{\vskip 0.5ex}

     \bf Inference    & Llama2-7B & \multirow{3}{*}{\cmark}   & \multirow{3}{*}{\xmark} &  \multirow{3}{*}{\parbox{7.5cm}{Inference efficiency poses a substantial challenge in LLMs, with a 100-fold delay compared to Enc2Dec models in our experiments.}}  \\  

     \bf Efficiency & WMT23 datasets & & &  \\ 
     & &  & & \\ \midrule

     \bf Pretraining Resource  & Llama2/MLLMs-7B & \multirow{3}{*}{-} & \multirow{3}{*}{\xmark} & \multirow{3}{*}{\parbox{7.5cm}{LLMs exhibit suboptimal performance in low-resource languages, stemming from the imbalance in pretraining resources.}}  \\

     \bf Imbalance & WMT23 datasets & & &  \\ 
     & & & & \\ \cdashline{1-5}\noalign{\vskip 0.5ex}

     \bf Evaluation Issues & Llama2-7B & \multirow{3}{*}{-} & \multirow{3}{*}{\xmark} &  \multirow{3}{*}{\parbox{7.5cm}{The issue of automatic evaluation has arisen due to the divergence between human and automated assessments of LLM translation outputs.}} \\
     
     & WMT23 datasets & & &  \\  
     & & & & \\ \bottomrule
        
    \end{tabular}
    }
    \caption{An overview of revisiting MT challenges in the context of LLMs. The first six lines discuss the classic MT challenges, while the last two lines focus on specific challenges that arise in LLM scenarios.
    The ``\cmark'' symbolizes largely addressed issues, while the ``\xmark'' represents persisting unresolved challenges.
    }
    \label{tab:overview}
\end{table*}



In the Natural Language Processing community, one of the most critical and longstanding tasks is Machine Translation (MT), which aims to convert human languages from one form to another \cite{koehn2009statistical,poibeau2017machine}. 
As the demand for effective translation systems continues to grow, researchers have been striving to develop models that can tackle the inherent challenges of this complex task. 
In this context, the six challenges about MT proposed by \newcite{koehn-knowles-2017-six} have been widely recognized and studied by numerous studies, with many efforts revolving around them \cite{chu-wang-2018-survey,neishi-yoshinaga-2019-relation,garg-etal-2019-jointly,pang2023rethinking}.

The emerging Large Language Models (LLMs) have been a significant breakthrough in NLP \cite{touvron2023llama,openai2023gpt4,touvron2023llama2}.
LLMs have demonstrated remarkable capabilities, outperforming traditional approaches and setting new benchmark performance for various applications such as machine translation \cite{lyu2023new,zhu2023multilingual,zhang2023prompting,wang2023document}. 
LLMs exhibit remarkable translation capabilities for major languages, owing to their extensive pretraining on vast amounts of unpaired data. This implies a significant advancement over conventional techniques.
Consequently, recent studies have employed LLMs for translation tasks \cite{jiao2023parrot,alves-etal-2023-steering}, achieving remarkable performance.
However, it is unclear \textit{how LLMs fare against the six classical challenges,}
This intriguing question warrants further investigation and discussion.

To gain insights into LLM-based MT research and identify paths for advancement,
we train Llama2 models as German-to-English translation systems and evaluate their abilities in addressing the six classic challenges.
Note that English is a high-resource language in the Llama2 pretraining data \cite{touvron2023llama2} and German is a relatively high-resource language, ensuring the model's competence in these languages.
Furthermore, we identify two LLM-specific challenges: pretraining resource imbalance and human-like evaluation issues.
German, Chinese, Ukrainian, and Hebrew, are included for English-to-X low-resource translation tasks to assess the effects of resource imbalance.
Table~\ref{tab:overview} summarizes our key findings, revealing that LLMs have successfully tackled data quantity and long sentence translation challenges but still face unresolved issues.

\section{Experimental Setup}



\subsection{Large Language Models}
\label{sec:llm_settings}


This paper focuses on decoder-only LLMs, a popular architecture in recent years \cite{openai2023gpt4}. 
Open-source LLMs are typically trained on datasets dominated by a single major language \cite{touvron2023llama2, yang2024qwen2}. 
Pretrained Llama2 models are English-centric and highly representative in the LLM community, which have only undergone pre-training without extensive fine-tuning \cite{touvron2023llama2}.
Therefore, we select pretrained Llama2 models as our base models for most experiments on classic machine translation challenges in Section~\ref{sec:classic}. 
This choice ensures controlled, reliable comparisons and aligns with current research, enhancing the credibility of our findings.
We employ the Llama2-7B and Llama2-13B models, which are open-source language models with 7 billion and 13 billion parameters, respectively \cite{touvron2023llama2}.
For fine-tuning, we train the model using the Alpaca dataset to enhance its instruction-following capabilities. 
Below are two training settings based on the input format of paired data.

\begin{itemize}[leftmargin=*,topsep=0.1em,itemsep=0.1em,parsep=0.1em]
    \item \textbf{LLM-SFT} undergoes supervised fine-tuning (SFT) using bilingual pairs in conjunction with the Alpaca dataset \cite{alpaca}, where the bilingual pairs adopt the Alpaca format.
    \item  \textbf{LLM-CPT-SFT} involves continuous pretraining (CPT) of the Llama2 model on concatenated translation pairs \cite{zhu2023multilingual}, followed by fine-tuning using the Alpaca dataset.
\end{itemize}

Unless stated otherwise, LLM-SFT and LLM-CPT-SFT refer to 7-billion-parameter models. 

In Section~\ref{sec:mllms}, we explore the pretraining resource imbalance, a challenge specific to the multilingual translation of LLMs. To address this, we compare two existing multilingual LLMs, ALMA-7B \cite{xu2024a} and TowerInstruct-7B \cite{tower_llm_2024}, as baselines. These models are only evaluated in this section since the focus here is multilingual translation, contrasting with the bilingual nature of the majority of our Llama2-based experiments in Section~\ref{sec:c2}.


\subsection{Small Encoder-to-Decoder Models}


Encoder-to-decoder (Enc2Dec) models are widely recognized as the most effective framework for translation tasks \cite{vaswani2023attention}. 
We utilize the Fairseq\footnote{\url{https://github.com/facebookresearch/fairseq}} toolkit to train Enc2Dec models adhering to the model architectures proposed by \newcite{vaswani2023attention}. 
Specifically, the base architecture is employed for training models on small bilingual datasets with $500$k pairs or fewer, while the large architecture is used for datasets comprising $1$M pairs or more \cite{pang2023rethinking}.
For each language pair, we use SUBNMT toolkit\footnote{\url{https://github.com/rsennrich/subword-nmt}} to learn byte-pair encoding codes and transform words into subwords \cite{sennrich-etal-2016-edinburgh,sennrich-etal-2016-neural}.

\subsection{Data Conditions}

We employ the German-to-English (De2En) parallel data (300 million (M)) procured from the WMT23 translation tasks.\footnote{\url{https://www2.statmt.org/wmt23/translation-task.html}}
Our methodology encompasses training a translation model through random sampling on the dataset, extracting up to $20$M translation pairs for each task.
To safeguard the robustness of our evaluation process, we utilize the most recent publicly accessible generaltest2023 from WMT23, thereby precluding potential data leakage.\footnote{\url{https://github.com/wmt-conference/wmt23-news-systems/tree/master/txt}}
To assess LLMs in multi-domain tasks, we conduct an experiment with the multi-domain dataset for German-to-English translation obtained from OPUS \cite{tiedemann-2012-parallel,aharoni-goldberg-2020-unsupervised}.

\subsection{Training and Evaluation}



\paragraph{Training} 
For LLMs, we train each model with a learning rate of $2e$-$5$, batch size of $48$, over 3 epochs on 32 A100 GPUs, saving checkpoints every 500 steps and reporting the best score.
For Enc2Dec models, we use early stopping with a patience of 20, halting training when validation performance plateaus or declines \cite{yao2007early}.
We use the Adam optimizer \cite{kingma2014adam} with $\beta_{1}=0.9$, $\beta_{2}=0.9$, $\beta_{2}=0.98$, and $\epsilon=10^{-9}$ to optimize Enc2Dec models.


\paragraph{Evaluation}
We employ two widely recognized automatic metrics: BLEU \cite{papineni-etal-2002-bleu}, using tokenized and case-sensitive SacreBLEU\footnote{\url{https://github.com/mjpost/sacrebleu}} \cite{post-2018-call}, and COMET-DA \cite{rei-etal-2020-comet}, based on the Unbabel/wmt22-comet-da model, a reference-based evaluation approach.

The BLEU metric assesses translation quality by measuring surface-level similarity through n-gram matching between the generated output and reference translations. While BLEU may not fully capture the semantic intricacies of the translation, it provides a reliable indication of fluency and lexical accuracy, ensuring that the basic grammatical structure of the output is maintained \cite{post-2018-call}.

To complement BLEU’s surface-level evaluation, we utilize COMET-DA to assess deeper semantic-level alignment between the translated output and reference text. 
COMET-DA leverages pre-trained language models to provide a more nuanced assessment of sentence-level translation adequacy and meaning preservation \cite{rei-etal-2020-comet}. 
By combining both metrics, we ensure a more comprehensive evaluation that accounts for both the syntactic and semantic quality.


\begin{table*}[!ht]
\small
  \centering
\begin{subtable}[h]{1.0\textwidth}
\centering
\resizebox{\textwidth}{!}{
  \begin{tabular}{lccccc}
    \toprule
    \bf System$\downarrow$ & \bf Law & \bf Medical & \bf IT & \bf Koran & \bf Subtitles \\ \midrule
    \multirow{-1.6}{*}{\bf All Data}
    & \DrawThreePercentageBar{64.3}{61.5}{56.4}
    & \DrawThreePercentageBar{61.5}{59.7}{51.4}
    & \DrawThreePercentageBar{50.6}{47.5}{41.7}
    & \DrawThreePercentageBar{20.3}{18.6}{20.1}
    & \DrawThreePercentageBar{28.8}{27.0}{26.8} \\ \midrule

    \multirow{-1.6}{*}{\bf Law}
    & \cellcolor[rgb]{0.85,0.85,0.85}\DrawThreePercentageBar{63.9}{62.0}{59.0}
    & \DrawThreePercentageBar{38.3}{36.1}{21.7}
    & \DrawThreePercentageBar{31.1}{29.6}{13.1}
    & \DrawThreePercentageBar{14.5}{12.0}{2.7}
    & \DrawThreePercentageBar{22.3}{20.4}{5.4} \\ \midrule

    \multirow{-1.6}{*}{\bf Medical}
    & \DrawThreePercentageBar{30.2}{28.5}{18.3}
    & \cellcolor[rgb]{0.85,0.85,0.85}\DrawThreePercentageBar{61.4}{59.3}{56.5}
    & \DrawThreePercentageBar{29.2}{31.2}{11.4}
    & \DrawThreePercentageBar{13.6}{11.8}{1.9}
    & \DrawThreePercentageBar{22.0}{19.9}{4.3} \\ \midrule

    \multirow{-1.6}{*}{\bf IT}
    & \DrawThreePercentageBar{32.8}{30.3}{9.6}
    & \DrawThreePercentageBar{38.5}{36.9}{14.9}
    & \cellcolor[rgb]{0.85,0.85,0.85}\DrawThreePercentageBar{51.0}{47.4}{43.0}
    & \DrawThreePercentageBar{14.4}{11.7}{2.8}
    & \DrawThreePercentageBar{24.2}{23.1}{8.6} \\ \midrule

    \multirow{-1.6}{*}{\bf Koran}
    & \DrawThreePercentageBar{23.1}{22.3}{0.2}
    & \DrawThreePercentageBar{27.5}{28.15}{0.1}
    & \DrawThreePercentageBar{19.2}{16.8}{0.2}
    & \cellcolor[rgb]{0.85,0.85,0.85}\DrawThreePercentageBar{20.6}{20.3}{15.9}
    & \DrawThreePercentageBar{10.8}{10.6}{0.5} \\ \midrule

    \multirow{-1.6}{*}{\bf Subtitles}
    & \DrawThreePercentageBar{28.9}{27.1}{5.5}
    & \DrawThreePercentageBar{33.4}{33.5}{7.9}
    & \DrawThreePercentageBar{26.7}{26.9}{8.5}
    & \DrawThreePercentageBar{13.2}{11.6}{6.4}
    & \cellcolor[rgb]{0.85,0.85,0.85}\DrawThreePercentageBar{28.9}{28.1}{27.3}
    \\ \bottomrule
  \end{tabular}
}
\caption{BLEU scores}
\label{tab:c1_multi_domain_bleu}
\end{subtable}

\begin{subtable}[h]{1.0\textwidth}
\centering
\resizebox{\textwidth}{!}{
  \begin{tabular}{lccccc}
    \toprule
    \bf System$\downarrow$ & \bf Law & \bf Medical & \bf IT & \bf Koran & \bf Subtitles \\ \midrule
    \multirow{-1.6}{*}{\bf All Data}
    & \cometDrawThreePercentageBar{88.2}{87.6}{86.3}
    & \cometDrawThreePercentageBar{86.1}{85.9}{85.1}
    & \cometDrawThreePercentageBar{87.8}{87.3}{85.1}
    & \cometDrawThreePercentageBar{72.2}{71.7}{70.4}
    & \cometDrawThreePercentageBar{78.4}{77.8}{76.5} \\ \midrule

    \multirow{-1.6}{*}{\bf Law}
    & \cellcolor[rgb]{0.85,0.85,0.85}\cometDrawThreePercentageBar{88.4}{88.0}{85.9}
    & \cometDrawThreePercentageBar{83.0}{82.6}{66.0}
    & \cometDrawThreePercentageBar{79.4}{78.9}{59.4}
    & \cometDrawThreePercentageBar{70.4}{68.8}{40.6}
    & \cometDrawThreePercentageBar{76.3}{75.0}{46.5} \\ \midrule

    \multirow{-1.6}{*}{\bf Medical}
    & \cometDrawThreePercentageBar{76.8}{76.3}{54.7}
    & \cellcolor[rgb]{0.85,0.85,0.85}\cometDrawThreePercentageBar{85.8}{85.7}{83.5}
    & \cometDrawThreePercentageBar{78.9}{78.4}{52.8}
    & \cometDrawThreePercentageBar{69.3}{67.7}{38.1}
    & \cometDrawThreePercentageBar{75.3}{74.2}{44.1} \\ \midrule

    \multirow{-1.6}{*}{\bf IT}
    & \cometDrawThreePercentageBar{81.5}{80.1}{48.0}
    & \cometDrawThreePercentageBar{82.6}{82.1}{50.7}
    & \cellcolor[rgb]{0.85,0.85,0.85}\cometDrawThreePercentageBar{88.1}{87.5}{82.5}
    & \cometDrawThreePercentageBar{70.1}{68.1}{39.1}
    & \cometDrawThreePercentageBar{76.8}{76.2}{50.6} \\ \midrule

    \multirow{-1.6}{*}{\bf Koran}
    & \cometDrawThreePercentageBar{74.0}{74.0}{33.7}
    & \cometDrawThreePercentageBar{77.3}{77.9}{32.0}
    & \cometDrawThreePercentageBar{71.1}{71.2}{37.3}
    & \cellcolor[rgb]{0.85,0.85,0.85}\cometDrawThreePercentageBar{72.5}{72.5}{58.3}
    & \cometDrawThreePercentageBar{67.4}{66.7}{41.4} \\ \midrule

    \multirow{-1.6}{*}{\bf Subtitles}
    & \cometDrawThreePercentageBar{79.5}{78.9}{47.4}
    & \cometDrawThreePercentageBar{81.1}{81.3}{54.1}
    & \cometDrawThreePercentageBar{77.8}{79.1}{57.2}
    & \cometDrawThreePercentageBar{69.3}{68.2}{51.7}
    & \cellcolor[rgb]{0.85,0.85,0.85}\cometDrawThreePercentageBar{78.4}{78.6}{74.1}
    \\ \bottomrule
  \end{tabular}
}
\caption{COMET-DA scores}
\label{tab:c1_multi_domain_comet}
\end{subtable}
  \caption{Translation quality of multi-domain German-to-English translation tasks, where the system is trained on one domain (rows) and tested on another domain (columns). The {\color{black}black}, {\color{red}red}, and {\color{blue}blue} bars refer to the LLM-SFT-13B, LLM-SFT, and Enc2Dec models, respectively. LLMs improve the in- and out-of-domain translation qualities but still suffer from the problem of domain mismatch.}
  
  \label{tab:c1_multi_domain}
\end{table*}

\begin{table*}[t]
    \centering
    \small
    \scalebox{0.98}{
    \begin{tabular}{l p{4.4cm} p{4cm} p{4.3cm}} 
    \toprule
         &  \bf 1: Medical & \bf 2: IT & \bf 3: Subtitles \\ \midrule
      \bf Src  & Die Pipetten müssen in der intakten Folienverpackung aufbewahrt werden. & Methode die Berechnung der Vorhersage & Du kannst ihr nicht helfen, außer dass du jetzt den Tatort untersuchst. \\ \cdashline{1-4}\noalign{\vskip 0.5ex}
      \bf Ref & Stored pipettes must be kept in the intact foil package. & Method to calculate forecast & You can't do anything but help her by working the crime scene. \\
      \midrule
      \bf All & Pipettes must be stored in the intact foil pouch. & Method to calculate the forecast & You can't help her except by canvassing the scene. \\ \midrule
      \bf Law & The pipettes must be stored in the intact \textcolor{red}{sheet} pack. & Method \textcolor{blue}{of calculation} of the forecast
 & You can't help her, except by examining the scene of the crime. \\ \cdashline{1-4}\noalign{\vskip 0.5ex}
      \bf Medical & The pipettes must be stored in the intact foil pouch. &  \textcolor{blue}{Prediction method}
 & You can’ t help her now, except by \textcolor{forestgreen}{examining the scene of the accident}. \\ \cdashline{1-4}\noalign{\vskip 0.5ex}
      \bf IT & The pipettes must be stored in the intact \textcolor{red}{slide} pack. & Method \textcolor{blue}{for calculating} the forecast
  & You can't help her, except by examining the crime scene. \\ \cdashline{1-4}\noalign{\vskip 0.5ex}
      \bf Koran & The pipettes must be kept in \textcolor{red}{sterile} packaging. & \textcolor{blue}{A method} to calculate the prediction.
 & If you do not, you will not be able to help her. \\ \cdashline{1-4}\noalign{\vskip 0.5ex}
      \bf Subtitles & The pipettes must be stored in intact \textcolor{red}{slide wraps}. &  Method \textcolor{blue}{for calculating} the forecast
 & You can't help her, except to process the crime scene.  \\ \bottomrule
    \end{tabular}
    }
    \caption{Test examples of German-to-English from three domains, which are translated by the LLM-SFT-7B trained on domains listed in the first column. The \textcolor{red}{red}, \textcolor{blue}{blue}, and \textcolor{forestgreen}{green} color indicates the terminology mismatch, style mismatch, and hallucination phenomena, respectively.}
    \label{tab:mdexamples}
\end{table*}

\section{The Six Classical Challenges}
\label{sec:classic}

\subsection{Domain Mismatch}

Domain mismatch has long been a formidable challenge in the field of MT \cite{wang2020go}.
Given the extensive pretraining of LLMs on diverse data, an intriguing question arises: \textit{does the vast knowledge encapsulated in LLMs mitigate the domain mismatch issue in translation tasks?}
To explore this, we finetune the LLM using domain-specific parallel data and evaluate its performance on both in-domain (ID) and out-of-domain (OOD) translation tasks. Table~\ref{tab:c1_multi_domain} presents the results.

\begin{itemize}[leftmargin=*,topsep=0.1em,itemsep=0.1em,parsep=0.1em]


\item \textbf{LLMs excel in-domain but face challenges with domain shifts.}
The results indicate that LLM-based translation systems perform exceptionally well on in-domain tasks, as reflected by both surface-level and semantic-level metrics. For instance, in the law domain, the LLM-SFT model achieves a notable BLEU score of 62.0 and a COMET-DA score of 88.0, outperforming Enc2Dec models by approximately 3.0 BLEU points and 4.9 COMET-DA points.
However, despite improvements in out-of-domain (OOD) translation over Enc2Dec models, LLMs still demonstrate significant performance degradation when encountering domain shifts. In the Koran-to-Law OOD translation task, this decline can be as severe as 40.0 BLEU points and 14.4 COMET-DA points.
A qualitative analysis is conducted in Table~\ref{tab:mdexamples}.

\item \textbf{Terminology Mismatch}. 
A common error in translation systems is the inability to produce accurate domain-specific terminology. 
For instance, in the medical domain, the term ``Folienverpackung'' has been inadequately translated as ``slide pack'', ``sterile packaging'', and ``slide wraps'' by the models finetuned with IT, Korean, and Subtitles bitexts, respectively.
\item \textbf{Style Mismatch}. 
The LLM fails to generate a hypothesis that accurately matches the out-of-domain style. In the second example, the reference in the IT domain is titled ``Method to calculate forecast'', while the medical translation system produces ``Prediction method''.
\item \textbf{Hallucination}. 
In the third example, the medical translation system erroneously translates the term ``Tatort'' as ``accident'' instead of ``crime'', where ``accident'' is a prevalent term in the medical domain. This translation error has been identified as an Input-conflicting Hallucination in the study by \newcite{zhang2023hallucination}.
    
\end{itemize}


Another concern arises regarding the effective management of domain expertise when a single LLM is tasked with handling multiple domains. Experimental results from the translation system trained on all domains show that, while the LLM performs consistently across various tasks, it falls short compared to domain-specific models. 
For example, in the Koran translation test, the LLM-SFT model trained on all data lags behind the domain-specific Koran LLM-SFT model by $1.7$ BLEU points and $0.8$ COMET-DA points.

\pgfplotsset{superb legend/.style={legend style                                = {draw=none,
                 legend columns                          = 3,
                 /tikz/every even column/.append style   = {column sep=0.5cm,
                 text width=7em},
                 /tikz/every odd column/.append style    = {column sep=0.15cm,
                  text width=7em},
                 }}}

\begin{figure*}[!ht]
\centering
\begin{subfigure}{0.49\linewidth}
\begin{tikzpicture}
\pgfplotsset{
every axis legend/.append style={at={(1.13,1.03)},anchor=south},
}
\begin{axis}[
    legend columns=3,legend style={font=\small},
    xlabel={\bf Sizes of Parallel Data},
    ylabel={\bf BLEU},
    xmin=-0.2, xmax=8.2,
    ymin=2, ymax=50,
    ytick={5,10,15,20,25,30,35,40,45},
    xticklabels={0, 10k,50k,100k,500k,1M,5M,10M,20M},
    xtick={0,1,2,3,4,5,6,7,8},
    ymajorgrids=true,
    grid style=dashed,
    height=2.5in,
    width=1.0\textwidth,
    nodes near coords
]

\addplot[
    color=red,
    mark=triangle,
    mark options={scale=1.5}
    ]
    coordinates {
    (1,4.6)(2,10.4)(3,14.4)(4,23.3)(5,26.6)(6,33.7)(7,35.9)(8,35.8)
    };
    \addlegendentry{Enc2Dec}

\addplot[
    color=blue,
    mark=square,
    ]
    coordinates {
    (0,38.2)(1,40.2)(2,41.1)(3,41.6)(4,41.4)(5,40.9)(6,40.0)(7,38.9)(8,39.2)
    };
    \addlegendentry{LLM-SFT}

\addplot[
    color=forestgreen,
    mark=x,
    nodes near coords style={yshift=-15pt},
    mark options={scale=1.5}
    ]
    coordinates {
    (0,38.2)(1,38.1)(2,38.5)(3,39.0)(4,40.1)(5,39.2)(6,40.3)(7,38.4)(8,38.5)
    };
    \addlegendentry{LLM-CPT-SFT}
    
\end{axis}
\end{tikzpicture}
\end{subfigure}
\centering
\begin{subfigure}{0.49\linewidth}
\begin{tikzpicture}
\pgfplotsset{
every axis legend/.append style={at={(0.5,1.03)},anchor=south},
}
\begin{axis}[
    legend columns=3,legend style={font=\small},
    xlabel={\bf Sizes of Parallel Data},
    ylabel={\bf COMET-DA},
    xmin=-0.2, xmax=8.2,
    ymin=30, ymax=100,
    ytick={30,40,50,60,70,80,90},
    xticklabels={0, 10k,50k,100k,500k,1M,5M,10M,20M},
    xtick={0,1,2,3,4,5,6,7,8},
    ymajorgrids=true,
    grid style=dashed,
    height=2.5in,
    width=1.0\textwidth,
    nodes near coords
]

\addplot[
    color=red,
    mark=triangle,
    mark options={scale=1.5}
    ]
    coordinates {
    (1,37.2)(2,52.7)(3,60.7)(4,71.4)(5,73.6)(6,73.7)(7,81.0)(8,81.3)
    };

\addplot[
    color=blue,
    mark=square,
    ]
    coordinates {
    (0,82.6)(1,83.7)(2,83.8)(3,83.9)(4,83.8)(5,83.7)(6,83.5)(7,83.3)(8,83.3)
    };

\addplot[
    color=forestgreen,
    mark=x,
    nodes near coords style={yshift=-15pt},
    mark options={scale=1.5}
    ]
    coordinates {
    (0,82.6)(1,82.7)(2,82.9)(3,82.9)(4,82.8)(5,82.8)(6,82.3)(7,82.8)(8,82.9)
    };
    
\end{axis}
\end{tikzpicture}
\end{subfigure}

\caption{BLEU and COMET-DA scores for German-to-English systems, with ``0'' on the x-axis indicating models trained exclusively on the Alpaca dataset. LLMs reduce reliance on extensive parallel data.}
\label{fig:amountparalleldata} 

\end{figure*}
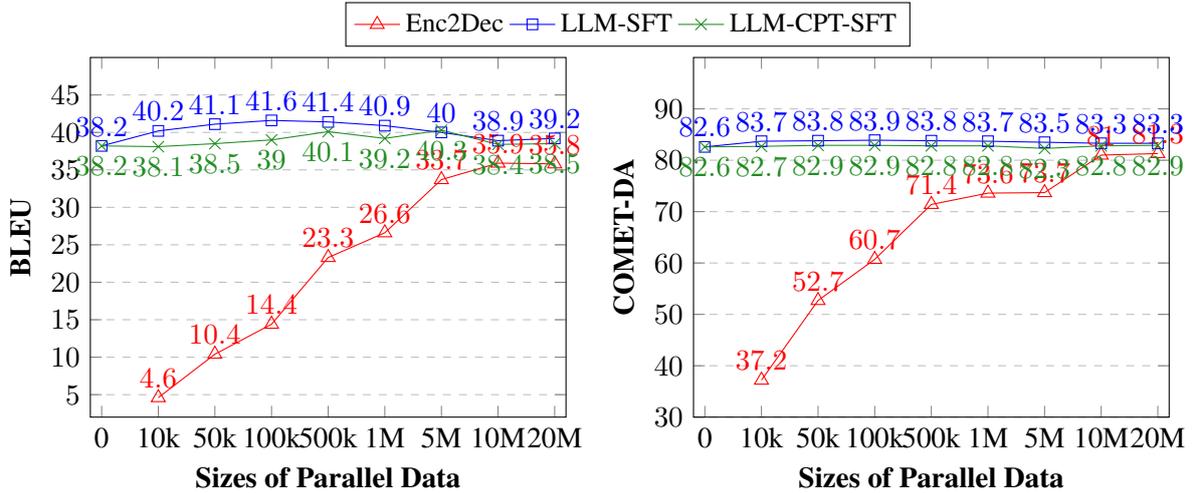
\pgfplotsset{superb legend/.style={legend style                                = {draw=none,
                 legend columns                          = 3,
                 /tikz/every even column/.append style   = {column sep=0.5cm,
                 text width=7em},
                 /tikz/every odd column/.append style    = {column sep=0.15cm,
                  text width=7em},
                 }}}

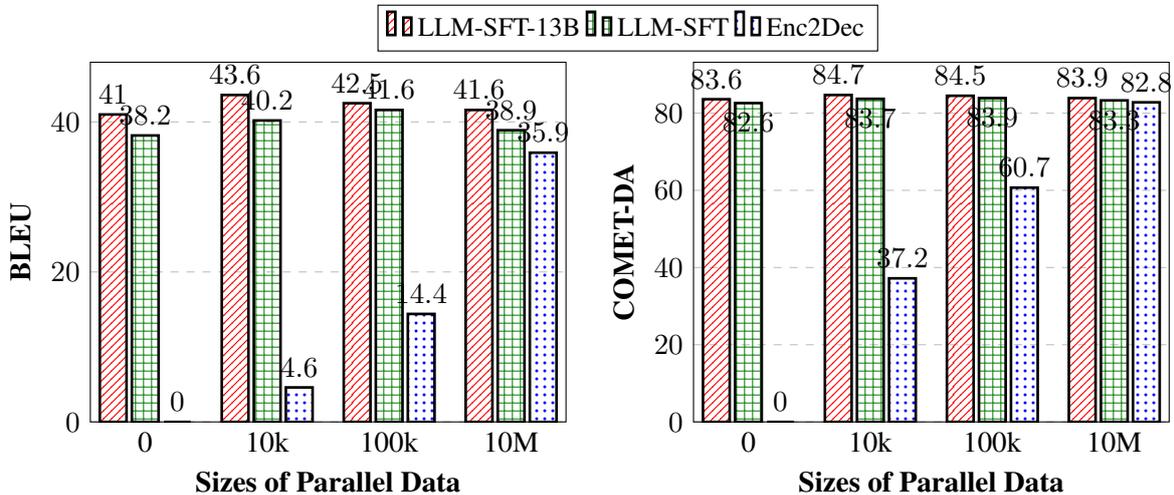
\begin{figure*}[!ht]
\centering
\begin{subfigure}{0.49\linewidth}
\begin{tikzpicture}
\pgfplotsset{
every axis legend/.append style={at={(1.13,1.03)},anchor=south},
}
    \begin{axis}[
        ybar,
        legend columns=3,legend style={font=\small},
        enlarge x limits=0.15,
        ylabel={\bf BLEU},
        xlabel={\bf Sizes of Parallel Data},
        symbolic x coords={0, 10k, 100k, 10M},
        xtick=data,
        ymajorgrids=true,
        x tick label style={yshift=0.15cm},
        grid style=dashed,
        nodes near coords,
        nodes near coords align={vertical},
        width=1.0\textwidth,
        height=2.5in,
        tick style={draw=none},
        legend cell align={left},
        ymin=0,
        ]

    \addplot [pattern=north east lines, pattern color=red, thick,line width=1pt] coordinates {(0, 41.0)(10k,43.6) (100k,42.5) (10M,41.6) };
    \addplot [pattern=grid, pattern color=forestgreen, thick,line width=1pt] coordinates { (0, 38.2)(10k,40.2) (100k,41.6) (10M,38.9)};
    \addplot [pattern=dots, pattern color=blue, thick,line width=1pt] coordinates { (0, 0)(10k,4.6) (100k,14.4) (10M,35.9)};
    \legend{LLM-SFT-13B, LLM-SFT, Enc2Dec}
    
    \end{axis}

\end{tikzpicture}
\end{subfigure}
\centering
\begin{subfigure}{0.49\linewidth}
\begin{tikzpicture}
\pgfplotsset{
every axis legend/.append style={at={(0.5,1.03)},anchor=south},
}
    \begin{axis}[
        ybar,
        legend columns=3,legend style={font=\small},
        enlarge x limits=0.15,
        ylabel={\bf COMET-DA},
        xlabel={\bf Sizes of Parallel Data},
        symbolic x coords={0, 10k, 100k, 10M},
        xtick=data,
        ymajorgrids=true,
        x tick label style={yshift=0.15cm},
        grid style=dashed,
        nodes near coords,
        nodes near coords align={vertical},
        width=1.0\textwidth,
        height=2.5in,
        tick style={draw=none},
        legend cell align={left},
        ymin=0,
        ]

    \addplot [pattern=north east lines, pattern color=red, thick,line width=1pt] coordinates {(0, 83.6)(10k,84.7) (100k,84.5) (10M,83.9) };
    \addplot [pattern=grid, pattern color=forestgreen, thick,line width=1pt, nodes near coords style={yshift=-15pt}] coordinates { (0, 82.6)(10k,83.7) (100k,83.9) (10M,83.3)};
    \addplot [pattern=dots, pattern color=blue, thick,line width=1pt] coordinates { (0, 0)(10k,37.2) (100k,60.7) (10M,82.8)};
    
    \end{axis}

\end{tikzpicture}
\end{subfigure}

    \caption{BLEU and COMET-DA scores for German-to-English systems. Increased parallel data adversely affects the performance of the Llama2-13B model.}
    \label{fig:c2_13B}
\end{figure*}

\subsubsection{Scaling Up to Llama2-13B}





We utilize the Llama2-13B model to evaluate whether the domain mismatch issue diminishes as the model size increases (Table~\ref{tab:c1_multi_domain}). The results reveal several important trends in multi-domain translation tasks.

\textbf{First, although the domain mismatch persists, reflected in the performance gap between in-domain and out-of-domain tasks, the 13B model trained on domain-specific data outperforms both the 7B model and Enc2Dec models.}
It achieves superior results on both sentence-level and semantic-level metrics. For instance, in the law domain, the 13B model attains a BLEU score of 63.9 and a COMET-DA score of 88.4, surpassing the 7B model by 1.9 BLEU points and 0.4 COMET-DA points. This suggests that stronger models can reduce, though not entirely eliminate, the domain mismatch problem. These findings raise the question of whether a more powerful LLM could fully resolve this issue.

\textbf{Second, the 13B model trained on all data consistently outperforms models trained on specific domains.} It falls short by only 0.3 BLEU and COMET-DA points compared to the 13B model trained on Koran data but demonstrates superior translation performance across all other domains. 
This indicates that a larger model can more effectively learn from multi-domain data due to its increased capacity.

In summary, increasing model capacity enhances overall performance across domains, though the domain mismatch issue remains a significant challenge.

\subsection{Amount of Parallel Data}
\label{sec:c2}

Parallel data is crucial for training encoder-to-decoder translation systems. 
With the emergence of LLMs, even a small corpus of high-quality parallel data can enhance their translation abilities \cite{jiao2023parrot}.
In this study, using the German-to-English translation task, we examine the impact of varying parallel data sizes, from $10$k to $20$M, on LLMs and assess the effectiveness of two training strategies for adapting LLMs into translation systems.
Our findings suggest:

\begin{itemize}[leftmargin=*,topsep=0.1em,itemsep=0.1em,parsep=0.1em]

    \item \textbf{A small amount of parallel data enhances LLM translation performance.} The LLM-SFT curve in Figure~\ref{fig:amountparalleldata} shows that supervised fine-tuning with 10k parallel data improves the BLEU score by $2$ and the COMET-DA score by $1.1$ compared to the Alpaca-only trained model. 
    Moreover, the LLM-SFT model trained 100k parallel data achieves the top BLEU score of $41.6$ and the top COMET-DA score of $83.9$.

    \item \textbf{An increasing amount of parallel data may degrade LLM translation performance.} Contrary to the belief that more parallel data improves translation quality, LLM models exhibit contrary results. 
    For both LLM-SFT and LLM-CPT-SFT, using large amounts of parallel data (e.g., 5M and 10M) negatively affects performance. Prior research suggests that LLMs acquire most knowledge during the pretraining stage \cite{zhou2023lima}., where unintentional exposure to bilingual signals occurs \cite{briakou-etal-2023-searching}. Excessive parallel data might disrupt this acquired knowledge.

    \item \textbf{Supervised fine-tuning outperforms continuous pretraining for utilizing additional parallel data.} Performance curves in Figure~\ref{fig:amountparalleldata} reveal that LLM-SFT consistently achieves better results, with a significant increase of up to $2.6$ BLEU scores and $1.0$ COMET-DA scores in the $100$k scenario.
    
\end{itemize}

In this section, we evaluate LLM translation systems with varying parallel data amounts. Our findings reveal that LLMs do not require extensive translation pair training, unlike conventional Enc2Dec models. 
This insight encourages researchers to explore efficient ways to utilize parallel data for LLM translation system enhancement, providing a potential direction for future studies to optimize bilingual knowledge and improve machine translation performance using LLMs.

\subsubsection{Scaling Up to Llama2-13B}

In this study, we utilize a stronger LLM, Llama2-13B, to delve deeper into the influence of parallel data on translation quality. We perform supervised fine-tuning on the parallel data, choosing 10k, 100k, and 10M parallel data as our evaluation points. A noteworthy trend emerges for stronger LLMs, as described below.


\textbf{For more advanced LLMs, such as Llama2-13B, the degradation point occurs earlier as the volume of parallel data increases.}
As illustrated in Figure~\ref{fig:c2_13B}, the LLM-SFT-13B model achieves its highest BLEU score of 43.6 and a COMET-DA score of 84.7 when trained on 10k parallel data. 
However, as the amount of parallel data increases to 100k and 10M, the model's performance declines, with BLEU scores falling to 42.5 and 40.0, respectively.
This finding suggests that LLMs require only a small amount of parallel data to enhance their translation capabilities for major pretrained languages. 
Conversely, stronger LLMs may be sensitive to performance degradation when exposed to increasing volumes of parallel data.

\input{tablesandfigures/c3_rarewords_new}
\begin{table*}[h!]
\centering
\begin{tabular}{llcccccc}
\toprule
  \multirow{2}{*}{\bf Freq. Bin}    & \multirow{2}{*}{\bf Counts}         & \multicolumn{2}{c}{\bf LLM-SFT-13B} & \multicolumn{2}{c}{\bf LLM-SFT} & \multicolumn{2}{c}{\bf Enc2Dec} \\ \cmidrule{3-4} \cmidrule{5-6} \cmidrule{7-8}
 & & \bf ACC(\%)          &  \bf Delete(\%)         & \bf ACC(\%)              &  \bf Delete(\%)            & 
 \bf ACC(\%)            & \bf Delete(\%)        \\ \midrule

0     & 80   & 42.27 & 12.01 & 48.84 & 6.57  & \textbf{50.17} & 8.25  \\
1     & 22   & 42.95 & 15.38 & \textbf{55.13} & 3.85  & 48.00 & 14.00 \\
2     & 11   & 11.54 & 7.69  & 23.72 & 0.00  & \textbf{28.47} & 0.00  \\
4     & 28   & 26.28 & 9.62  & 35.26 & 13.46 & \textbf{42.03} & 8.70  \\
8     & 22   & 23.19 & 8.70  & \textbf{36.96} & 8.70  & 34.09 & 9.09  \\
16    & 46   & 38.54 & 14.58 & 50.00 & 6.25  & \textbf{62.59} & 4.44  \\ \cdashline{1-8}\noalign{\vskip 0.5ex}
32    & 62   & 52.98 & 6.76  & \textbf{55.46} & 6.76  & 52.20 & 16.86 \\
64    & 82   & 53.88 & 8.52  & \textbf{56.44} & 3.98  & 50.98 & 18.28 \\
128   & 139  & \textbf{59.61} & 5.67  & 58.42 & 6.00  & 57.54 & 10.81 \\
256   & 165  & \textbf{56.38} & 8.47  & 56.28 & 8.20  & 56.25 & 14.66 \\
512   & 183  & 60.18 & 9.01  & \textbf{62.19} & 6.22  & 57.92 & 14.15 \\
999   & 244  & \textbf{62.30} & 8.37  & 62.06 & 5.62  & 56.39 & 12.48 \\
1999  & 295  & 63.21 & 6.46  & \textbf{65.05} & 6.40  & 61.20 & 10.38 \\
3999  & 380  & 66.33 & 4.11  & \textbf{67.85} & 4.83  & 60.30 & 10.13 \\
7999  & 425  & \textbf{66.28} & 3.89  & 64.10 & 3.56  & 61.34 & 8.80  \\
15999 & 554  & \textbf{65.31} & 4.34  & 63.51 & 4.91  & 57.62 & 9.80  \\
31999 & 609  & \textbf{65.00} & 3.77  & 64.13 & 4.88  & 58.48 & 11.98 \\
63999 & 680  & 65.05 & 3.75  & \textbf{65.66} & 3.97  & 59.72 & 10.66 \\
64000 & 2658 & 67.29 & 4.66  & \textbf{67.52} & 4.59  & 61.07 & 12.09 \\

\bottomrule

\end{tabular}
\caption{Precision of translation and delete rates concerning source word types with varying frequencies. ``Freq. Bin'' and ``Counts'' indicate the frequency upper bounds and the count of each source type word. ``ACC'' and ``Delete'' indicate the precision and deletion rates of word prediction.}
\label{tab:c313Brarewords}
\end{table*}

\subsection{Rare Word Prediction}
\label{sec:rare_word}

Rare word prediction is crucial in translation, especially for proper nouns, compounds, names, and loanwords referring to specific entities or items \cite{luong-etal-2015-addressing,sennrich-etal-2016-neural,wu2016googles}. Following previous approaches \cite{koehn-haddow-2012-interpolated,koehn-knowles-2017-six}, we evaluate the precision and deletion rate of rare words in both the LLM and Enc2Dec translation systems. Our analysis includes the best LLM-SFT system trained on $100$k language pairs and the top-performing Enc2Dec model trained on $10$ million language pairs, as discussed in Section~\ref{sec:c2}. 
Results in Figure~\ref{fig:rarewords} reveal that:

\begin{itemize}[leftmargin=*,topsep=0.1em,itemsep=0.1em,parsep=0.1em]

\item \textbf{The LLM translation model demonstrates higher precision and lower deletion rates for frequent words.}
Although Enc2Dec models are typically trained on extensive WMT datasets and expected to excel in generating precise in-domain words, the LLM-SFT model outperforms this expectation. The LLM model consistently achieves higher precision for words with a frequency above 16 compared to the Enc2Dec model. Specifically, for words in the frequency bin of (1999,3999], LLM-SFT achieves 67.85\% prediction accuracy while that of Enc2Dec is 60.30\%.
Besides, the deletion rates are consistently lower than those of the Enc2Dec model.

\item \textbf{The LLM translation model struggles with infrequent word prediction, leading to translation omissions.}
The LLM-SFT system exhibits low precision for words occurring less than $8$ times and high deletion rates. For words in the $(2,4]$ frequency bin, the LLM-SFT reports a word precision of $35.26\%$, while Enc2Dec achieves $42.03\%$. 
Moreover, the LLM deletion rate for the frequency bin $(2,4]$ is $13.46\%$, which is $4.76\%$ higher than the Enc2Dec model's $8.70\%$. 
Qualitatively, the LLM-SFT fails to translate compound rare words such as "blätterlos" (meaning "a plant has no leaves") and "lotrechtes" (meaning "perpendicular"), whereas Enc2Dec successfully generates "leaflessly" as a substitute for "blätterlos".
This finding suggests that LLMs struggle with the semantic and morphological complexities of rare or compound words.

\end{itemize}

Overall, our results show that LLMs achieve reliable word-level translation accuracy, but predicting rare words remains a major challenge that warrants further research.

\pgfplotsset{superb legend/.style={legend style                                = {draw=none,
                 legend columns                          = 3,
                 /tikz/every even column/.append style   = {column sep=0.5cm,
                 text width=7em},
                 /tikz/every odd column/.append style    = {column sep=0.15cm,
                  text width=7em},
                 }}}

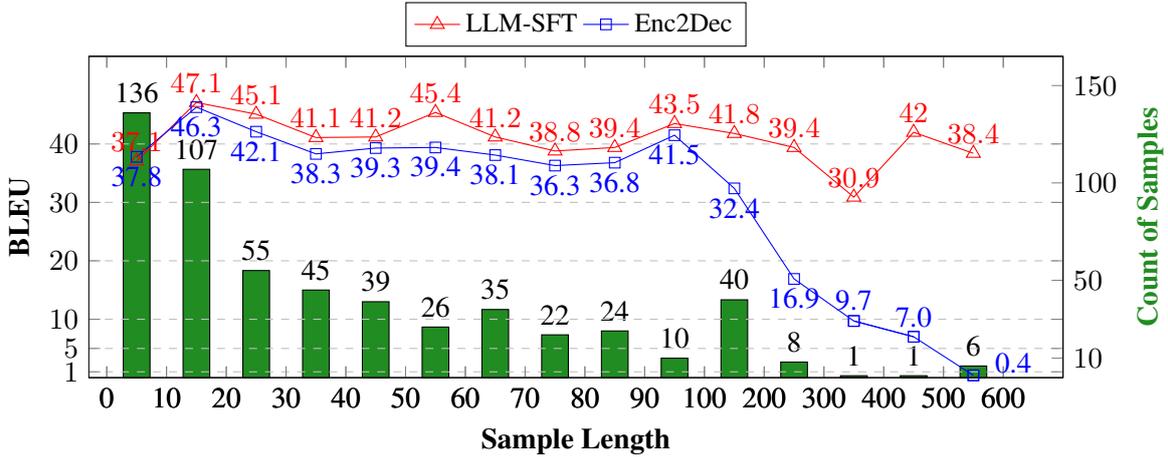
\begin{figure*}[ht]

    \centering

\begin{tikzpicture}
\pgfplotsset{
every axis legend/.append style={at={(0.5,1.03)},anchor=south},
}

\begin{axis}[
    legend columns=3,legend style={font=\small}, label style={font=\normalsize},
    axis y line*=right,
    ylabel={\bf {\color{forestgreen}Count of Samples}},
    xmin=-0.3, xmax=16,
    ymin=0, ymax=165,
    ytick={10,50,100,150},
    xtick={0,1,2,3,4,5,6,7,8,9,10,11,12,13,14,15},
    xticklabels={0,10,20,30,40,50,60,70,80,90,100,200,300,400,500,600},
    grid style=dashed,
    height=2.3in,
    width=0.9\textwidth,
    nodes near coords,
    point meta=explicit symbolic
]

\addplot[ybar, fill=forestgreen] 
    coordinates {
    (0.5,136)[136](1.5,107)[107](2.5,55)[55](3.5,45)[45](4.5,39)[39](5.5,26)[26](6.5,35)[35](7.5,22)[22](8.5,24)[24](9.5,10)[10](10.5,40)[40](11.5,8)[8](12.5,1)[1](13.5,1)[1](14.5,6)[6]
    };

\end{axis}

\begin{axis}[
    legend columns=3,legend style={font=\small},
    xlabel={\bf Sample Length},
    ylabel={\bf BLEU},
    xmin=-0.3, xmax=16,
    ymin=0, ymax=55,
    ytick={1,5,10,20,30,40},
    xtick={0,1,2,3,4,5,6,7,8,9,10,11,12,13,14,15},
    xticklabels={0,10,20,30,40,50,60,70,80,90,100,200,300,400,500,600},
    ymajorgrids=true,
    grid style=dashed,
    height=2.3in,
    width=0.9\textwidth,
    nodes near coords
]

\addplot[
    color=red,
    mark=triangle,
    mark options={scale=1.5}
    ]
    coordinates {
    (0.5,37.1)(1.5,47.1)(2.5,45.1)(3.5,41.1)(4.5,41.2)(5.5,45.4)(6.5,41.2)(7.5,38.8)(8.5,39.4)(9.5,43.5)(10.5,41.8)(11.5,39.4)(12.5,30.9)(13.5,42.0)(14.5,38.4)
    };
    \addlegendentry{LLM-SFT}

\addplot[
    color=blue,
    mark=square,
    nodes near coords style={yshift=-15pt},
    point meta=explicit symbolic
    ]
    coordinates {
    (0.5,37.8)[37.8](1.5,46.3)[46.3](2.5,42.1)[42.1](3.5,38.3)[38.3](4.5,39.3)[39.3](5.5,39.4)[39.4](6.5,38.1)[38.1](7.5,36.3)[36.3](8.5,36.8)[36.8](9.5,41.5)[41.5](10.5,32.4)[32.4](11.5,16.9)[16.9](12.5,9.7)[](13.5,7.0)[](14.5,0.4)[]
    }; \addlegendentry{Enc2Dec}

\addplot[
    color=blue,
    mark=square,
    point meta=explicit symbolic,
    forget plot
    ]
    coordinates {
    (12.5,9.7)[9.7](13.5,7.0)[7.0]
    };

\addplot[
    color=blue,
    mark=square,
    nodes near coords style={xshift=15pt},
    nodes near coords style={yshift=-3pt},
    point meta=explicit symbolic,
    forget plot
    ]
    coordinates {
    (14.5,0.4)[0.4]
    }; 



\end{axis}

\end{tikzpicture}

\caption{BLEU scores for German-to-English MT systems with varying sample lengths. Sentence-level translation involves lengths below $90$ words, while document-level translation concerns longer samples. LLMs improve long-sentence translation and consistently excel in document-level tasks.}
\label{fig:c3-sentlen}

\end{figure*}


\input{tablesandfigures/c4_sentlevel_comet}

\subsubsection{Scaling Up to Llama2-13B}

To further examine the poor performance in predicting unknown and low-frequency words, we conducted an additional evaluation focused on rare word prediction using the Llama2-13B model trained on 10k German-to-English parallel data, as shown in Figure~\ref{fig:c2_13B}. For a clearer comparison, we also present detailed results for the LLM-SFT-7B and Enc2Dec models in Table~\ref{tab:c313Brarewords}. Based on our findings, we observe that:

\textbf{A stronger LLM, such as Llama2-13B, more evidently struggles with predicting rare words.} Despite its high BLEU score, the stronger model consistently underperforms compared to LLM-SFT for rare words occurring less than 16 times. Specifically, for words in the frequency bin (1,2], LLM-SFT-13B achieves a prediction accuracy of 11.54\%, while LLM-SFT-7B and Enc2Dec report accuracy scores of 23.72\% and 28.47\%, respectively. This finding highlights the challenge LLMs face in predicting rare words.

The challenge of accurately generating rare words in LLMs arises from their intrinsic properties and decoding strategies.
LLMs, trained using a causal language modeling objective, learn high-frequency words more effectively due to their prevalence in training data \cite{radford2018improving}.
In contrast, rare words appear less frequently, resulting in a lower level of understanding and mastery by the model.
Techniques like greedy decoding or beam search, employed during text generation, favor high-frequency words based on probability distribution \cite{Holtzman2020The}.
Accordingly, our experiments reveal that more powerful models tend to exacerbate the phenomenon of imprecise generation of rare words.

\subsection{Translation of Long Sentences}

The length of the source text poses a significant challenge for MT systems due to the need for accurate contextual capture \cite{wang-etal-2017-exploiting-cross,wang2023document}. 
Given LLMs' extended context window, particularly Llama2's $4,096$ maximum input length \cite{touvron2023llama2}, we investigate how LLMs handle long sentence translation.

In this section, we evaluate model performance across varying sentence lengths on the generaltest2023 test set, segmented following previous settings \cite{koehn-knowles-2017-six}. We categorize the test set into groups based on sentence length, from 1-9 words up to 80-89 words. For document-level translation assessment, we include groups with 100-199 words to 400-499 words, merging with the original documents for a final group of 500-599 words. The test set's longest document contains 582 words. 
The best models in Section~\ref{sec:c2} are adopted for evaluation, which are the LLM-SFT trained on $100$k parallel data and the Enc2Dec model trained on $20$M parallel data.
Figure~\ref{fig:c3-sentlen} illustrates the sentence and document count per group and presents the BLEU scores for each test.
Figure~\ref{fig:c3-sentlen-comet} presents the COMET-DA score for sentence-level translation cases.

\begin{itemize} [leftmargin=*,topsep=0.1em,itemsep=0.1em,parsep=0.1em]
    \item \textbf{The LLM translation system excels at translating long sentences.}
    Based on the score curves for BLEU and COMET-DA evaluations, the LLM-SFT model outperforms the Enc2Dec model in both surface-level and semantic-level translation tasks for sentences up to 80 words in length. 
    However, for sentences shorter than 10 words, the LLM-SFT model lags behind the Enc2Dec model by 0.7 BLEU points. 
    In contrast, in the semantic-level evaluation, the LLM-SFT model surpasses the Enc2Dec model by 0.5 COMET-DA points. 
    This suggests that the LLM-based model effectively preserves high-quality semantic information in translation tasks.

   \item \textbf{LLMs demonstrate superior document translation capabilities.} As shown in Figure~\ref {fig:c3-sentlen}, for sentences exceeding $100$ words, LLMs maintain consistent high performance, whereas the Enc2Dec model's curve steeply falls to a $0.4$ BLEU score. For documents with more than 500 words, LLM-SFT achieves a $38.4$ BLEU score, indicating a substantial difference and potential in document-level translation tasks.
    
\end{itemize}




The findings presented above empirically demonstrate the proficiency of LLMs in translating sentences of varying lengths and indicate their potential for document-level translation tasks. 
It is important to note that the LLM-SFT model utilized in Section~\ref{sec:c2} was trained solely on sentence-level parallel data. 
Despite this limitation, the LLM-SFT model exhibits strong performance for translation tasks involving up to 600 words, highlighting its advantages in handling long context windows and leveraging pretrained knowledge.

\begin{table}[t!]
    \centering
    \small
\resizebox{0.48\textwidth}{!}{
    \begin{tabular}{lc}
    \toprule
    \bf System & \bf German-to-English \\ \midrule
    \bf GPT4  \cite{openai2023gpt4}  & 39.5  \\ 
    \bf Llama-MT \cite{zefeng-llama-mt} & 39.7 \\ \midrule
    \bf LLM-SFT-0k     &  22.2 \\
    \bf LLM-SFT-100k & 36.3 \\
    \bottomrule
    \end{tabular}
    }
    \caption{d-BLEU scores for TED document-level translation tasks. Llama2-SFT-* are models trained with 0k and 100k German-to-English parallel data in Figure~\ref{fig:amountparalleldata}, respectively.}
    \label{tab:docmt}
\end{table}

\subsubsection{Document-Level Translation}



We further evaluate our models' document-level translation competencies by directly testing them on the TED German-to-English document-level translation tasks \cite{cettolo-etal-2017-overview}.\footnote{https://wit3.fbk.eu/2017-01-d} 
For reference, we include the GPT-4 and Llama-MT models \cite{zefeng-llama-mt} as baselines.

\textbf{LLMs fine-tuned on sentence-level parallel data demonstrate proficiency in document-level translation tasks.} 
The Llama2-SFT-100k model registers a d-BLEU score of 36.3, significantly surpassing the Llama2-SFT-0k model's score of 22.2. This model also exhibits competitive performance against the leading GPT-4 and Llama-MT models, with only slightly lower d-BLEU scores. These findings further validate the robustness of LLMs' pretrained knowledge for document-level translation tasks \cite{wang2023document}, highlighting their adaptability to more intricate, context-dependent translation scenarios.

\subsection{Word Alignment}

Previous studies extracted word alignment from the attention matrix within encoder-to-decoder translation models \cite{garg-etal-2019-jointly,chen-etal-2020-accurate,zenkel-etal-2020-end} and used it to interpret translation models \cite{yang-etal-2020-sub,yang2021context}.

\begin{figure}[t!]
    \centering
    \includegraphics[width=0.48\textwidth]{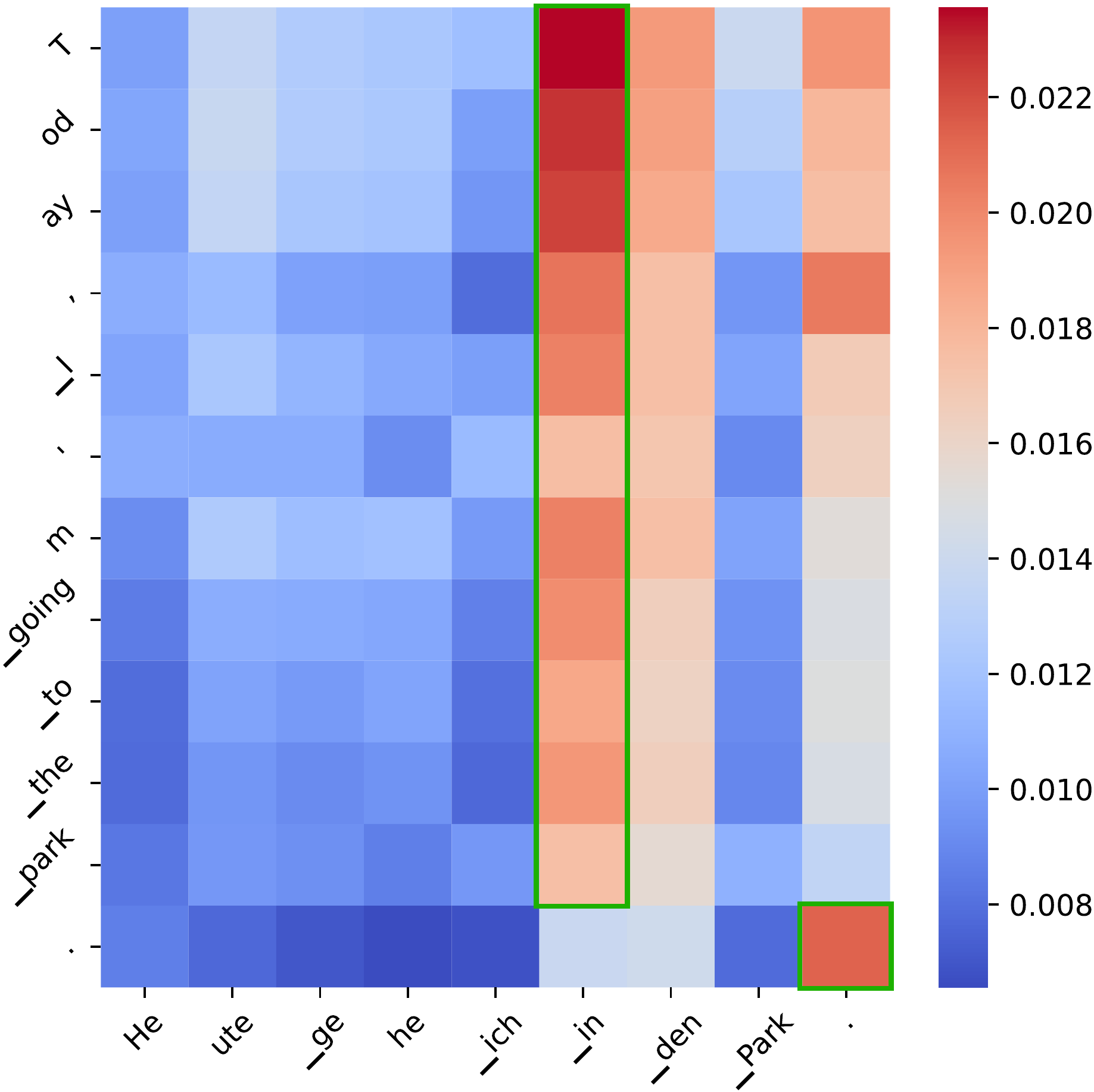}
    \caption{Average attention weight of target-to-source sentences of the LLM-SFT model. The left column is a target English sentence ``Today, I'm going to the park.'', and the bottom row is the source German sentence ``Heute gehe ich in den Park.'', tokenized by the Llama2 tokenizer.}
    \label{fig:attnweight}
\end{figure}

In this section, we explore two research questions: 1) \emph{Is it feasible to extract word alignment from LLM attention weights?} and 2) \emph{Can word alignment shed light on the LLM translation process?} To address these questions, we conduct a case study using the LLM-SFT model to process the instruction input and target sentence. We extract attention weights indicating the relationship between source and target words, as shown in Figure~\ref{fig:attnweight}. Our findings include:

\begin{itemize}[leftmargin=*,topsep=0.1em,itemsep=0.1em,parsep=0.1em]
    \item \textbf{Extracting alignments from LLM attention weights is not feasible.}
    Figure~\ref{fig:attnweight} displays the average attention weight across $32$ layers in the LLM-SFT translation model. Results reveal that each target sub-token tends to attend the same source token, in this case, ``$\_$in'', suggesting that attention weights do not explicitly provide word alignment information \cite{moradi-etal-2021-measuring}.

    \item \textbf{Aggregated attention weights offer clues for interpreting LLMs.}
    The observed phenomenon, where target tokens attend the same source tokens, aligns with \newcite{wang-etal-2023-label}'s findings. They discover that LLMs tend to aggregate information into one token, and predicted tokens pay the most attention to this token during inference, referred to as the anchor. 
\end{itemize}

To obtain word alignment, methods such as prediction difference \cite{li-etal-2019-word} and prompt design offer promising directions for further investigation. However, the most significant challenge lies in interpreting LLMs, for which the insights from this section provide valuable guidance for future research.

     


\pgfplotsset{superb legend/.style={legend style                                = {draw=none,
                 legend columns                          = 3,
                 /tikz/every even column/.append style   = {column sep=0.5cm,
                 text width=7em},
                 /tikz/every odd column/.append style    = {column sep=0.15cm,
                  text width=7em},
                 }}}

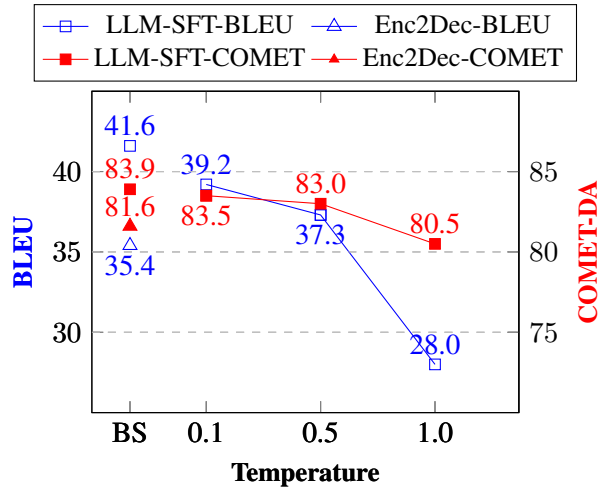
\begin{figure}[t]
    \centering

\begin{tikzpicture}[font=\normalsize]
\pgfplotsset{
every axis legend/.append style={at={(0.5,1.03)},anchor=south},
}
\begin{axis}[
    legend columns=2,legend style={font=\small},label style={font=\small},
    xlabel={\bf Temperature},
    ylabel={\bf {\color{blue}BLEU}},
    xmin=0, xmax=11.2,
    ymin=25, ymax=45,
    ytick={30,35,40},
    xticklabels={BS,0.1,0.5,1.0},
    xtick={1,3,6,9},
    ymajorgrids=true,
    grid style=dashed,
    height=2.3in,
    width=0.45\textwidth,
    nodes near coords,
    point meta=explicit symbolic
]

\addplot[
    color=blue,
    mark=square,
    ]
    coordinates {
    (-1,-1) 
    };
    \addlegendentry{LLM-SFT-BLEU}

\addplot[
    color=blue,
    mark=triangle,
    mark options={scale=1.5},
    nodes near coords style={yshift=-15pt},
    ]
    coordinates {
    (-1,-1) 
    };
    \addlegendentry{Enc2Dec-BLEU}

\addplot[
    color=red,
    mark=square*,
    ]
    coordinates {
    (-1,-1) 
    };
    \addlegendentry{LLM-SFT-COMET}

\addplot[
    color=red,
    mark=triangle*,
    ]
    coordinates {
    (-1,-1) 
    };
    \addlegendentry{Enc2Dec-COMET}


\end{axis}

\begin{axis}[
    legend columns=4,legend style={font=\small},label style={font=\small},
    xlabel={\bf Temperature},
    ylabel={\bf {\color{blue}BLEU}},
    xmin=0, xmax=11.2,
    ymin=25, ymax=45,
    ytick={30,35,40},
    xticklabels={BS,0.1,0.5,1.0},
    xtick={1,3,6,9},
    ymajorgrids=true,
    grid style=dashed,
    height=2.3in,
    width=0.45\textwidth,
    nodes near coords,
    point meta=explicit symbolic
]

\addplot[
    color=blue,
    mark=square,
    ]
    coordinates {
    (3,39.2) [39.2]  (6,37.3) (9,28.0) [28.0]
    };
    
\addplot[
    color=blue,
    mark=square,
    forget plot,
    nodes near coords style={yshift=-15pt},
    ]
    coordinates {
   (6,37.3) [37.3] 
    };

\addplot[
    color=blue,
    mark=square,
    forget plot,
    ]
    coordinates {
    (1,41.6) [41.6]
    };

\addplot[
    color=blue,
    mark=triangle,
    mark options={scale=1.5},
    nodes near coords style={yshift=-15pt},
    ]
    coordinates {
    (1,35.4) [35.4]
    };

\end{axis}

\begin{axis}[
    legend columns=4,legend style={font=\small}, label style={font=\small},
    axis y line*=right,
    ylabel={\bf {\color{red}COMET-DA}},
    xmin=0, xmax=11.2,
    ymin=70, ymax=90,
    ytick={75,80,85},,
    xticklabels={BS,0.1,0.5,1.0},
    xtick={1,3,6,9},
    ymajorgrids=true,
    grid style=dashed,
    height=2.3in,
    width=0.45\textwidth,
    nodes near coords,
    point meta=explicit symbolic
]

\addplot[
    color=red,
    mark=square*,
    ]
    coordinates {
    (3,83.5) [] (6,83.0) [83.0] (9,80.5) [80.5]
    };

\addplot[
    color=red,
    mark=square*,
    forget plot,
    nodes near coords style={yshift=-15pt},
    ]
    coordinates {
    (3,83.5) [83.5] 
    };


\addplot[
    color=red,
    mark=square*,
    forget plot,
    ]
    coordinates {
    (1,83.9) [83.9] 
    };

\addplot[
    color=red,
    mark=triangle*,
    mark options={scale=1.5},
    ]
    coordinates {
    (1,81.6) [81.6] 
    }; 

\end{axis}

\end{tikzpicture}

\caption{BLEU and COMET-DA scores of the German-to-English systems. ``BS'' indicates the beam search method with a beam size of $5$.} 
\label{fig:c6-bleucomet} 

\end{figure}

\subsection{Inference Efficiency}

In the realm of inference, two major concerns are inference strategies and efficiency. Beam search and sampling are commonly employed strategies. Beam search predicts the most promising word within a predefined beam size to generate a sentence \cite{freitag-al-onaizan-2017-beam}, while sampling randomly selects the next word based on the word probability distribution. 
Previous studies have examined the impact of beam size on beam search performance \cite{koehn-knowles-2017-six}, and practitioners in the machine translation field commonly use beam sizes of 4 or 5 \cite{vaswani2023attention}.
However, due to the extensive size of LLMs, inference efficiency becomes a more challenging issue, with recent works proposing to accelerate the inference process \cite{li-etal-2023-compressing,alizadeh2024llm}. 
In this section, we first analyze the performance difference between these two inference strategies, then discuss inference efficiency. Apart from BLEU, we include the COMET-DA evaluation \cite{rei-etal-2022-comet} for a semantic-level comparison. We set the beam size to $5$ for beam search and vary the temperature with $0.1$, $0.5$, and $1.0$ for sampling.

\begin{itemize}[leftmargin=*,topsep=0.1em,itemsep=0.1em,parsep=0.1em]

\item  \textbf{Sampling underperforms beam search in BLEU, but the difference is less pronounced in COMET-DA.}
Figure~\ref{fig:c6-bleucomet} shows that LLM-SFT achieves $41.6$ BLEU and $83.9$ COMET-DA scores with beam search inference. In contrast, using sampling with a temperature of $0.1$, LLM-SFT attains $39.2$ BLEU and $83.5$ COMET-DA scores. 
Our results indicate that LLM-SFT, using sampling with a temperature of $0.1$, achieves an accuracy of $55.79\%$ in predicting unknown words, but exhibits lower accuracies for frequent words compared to beam search. 
This suggests that the sampling method's ability to explore more diverse candidate sequences contributes to predicting rare words.

\item \textbf{Compared to Enc2Dec, inference latency poses a significant challenge in utilizing LLMs for MT.}
LLMs require an average of 30 seconds for inference, whereas Enc2Dec models require only 0.3 seconds on average, indicating a nearly 100-fold difference. 
This substantial discrepancy in inference latency between LLMs and Enc2Dec models presents a significant hurdle in the practical application of LLMs for machine translation. 
The longer inference time of LLMs can be attributed to their large model size and extensive parameters.

\end{itemize}

Current research is exploring ways to reduce the inference latency of LLMs, such as model compression or hardware acceleration \cite{dettmers2022llmint8, agrawal2023sarathi, frantar2023sparsegpt, dettmers2023spqr, lin2023awq, kim2023squeezellm}. Additionally, further exploration could optimize the performance of sampling methods in handling rare words, potentially enhancing the overall translation quality of LLMs.



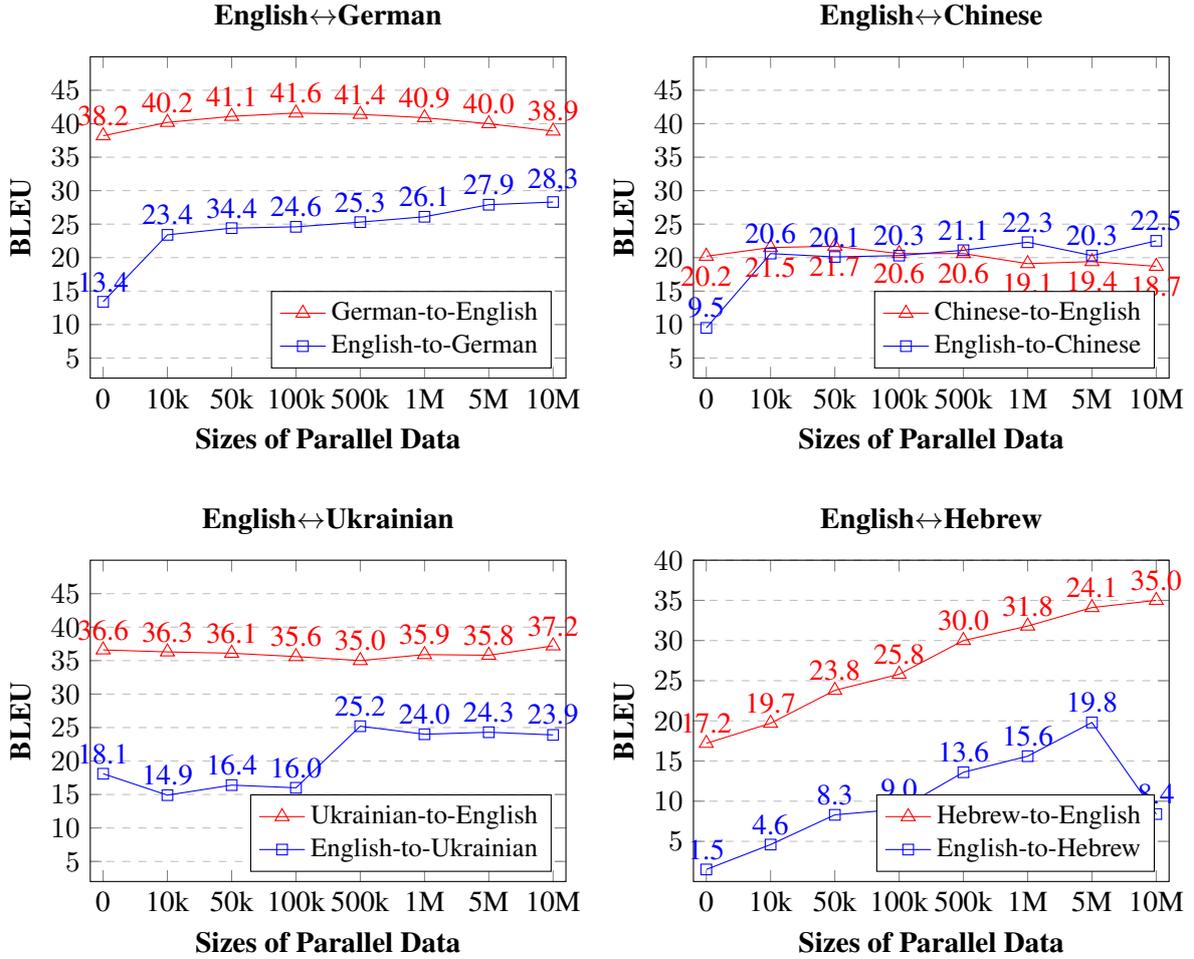
\begin{figure*}[th]
\centering
\begin{subfigure}[b]{0.49\textwidth}
\centering
\begin{tikzpicture}
\begin{axis}[
    legend columns=1,legend style={font=\small},
    title={\bf English$\leftrightarrow$German},
    xlabel={\bf Sizes of Parallel Data},
    ylabel={\bf BLEU},
    xmin=-0.2, xmax=7.2,
    ymin=2, ymax=50,
    ytick={5,10,15,20,25,30,35,40,45},
    xticklabels={0, 10k,50k,100k,500k,1M,5M,10M},
    xtick={0,1,2,3,4,5,6,7},
    legend pos=south east,
    ymajorgrids=true,
    grid style=dashed,
    height=2.3in,
    width=1.0\textwidth,
    nodes near coords,
    point meta=explicit symbolic
]

\addplot[
    color=red,
    mark=triangle,
    mark options={scale=1.5},
    nodes near coords style={font=\normalsize},
    ]
    coordinates {
    (0,38.2)[38.2](1,40.2)[40.2](2,41.1)[41.1](3,41.6)[41.6](4,41.4)[41.4](5,40.9)[40.9](6,40.0)[40.0](7,38.9)[38.9]
    };
    \addlegendentry{German-to-English}

\addplot[
    color=blue,
    mark=square,
    nodes near coords style={font=\normalsize},
    ]
    coordinates {
    (0,13.4)[13.4](1,23.4)[23.4](2,24.4)[34.4](3,24.6)[24.6](4,25.3)[25.3](5,26.1)[26.1](6,27.9)[27.9](7,28.3)[28,3]
    };
    \addlegendentry{English-to-German}

\end{axis}
\end{tikzpicture}
\end{subfigure}
\begin{subfigure}[b]{0.49\textwidth}
\centering
\begin{tikzpicture}

\begin{axis}[
    legend columns=1,legend style={font=\small},
    title={\bf English$\leftrightarrow$Chinese},
    xlabel={\bf Sizes of Parallel Data},
    ylabel={\bf BLEU},
    xmin=-0.2, xmax=7.2,
    ymin=2, ymax=50,
    ytick={5,10,15,20,25,30,35,40,45},
    xticklabels={0, 10k,50k,100k,500k,1M,5M,10M},
    xtick={0,1,2,3,4,5,6,7},
    legend pos=south east,
    ymajorgrids=true,
    grid style=dashed,
    height=2.3in,
    width=1.0\textwidth,
    nodes near coords,
    point meta=explicit symbolic
]

\addplot[
    color=red,
    mark=triangle,
    mark options={scale=1.5},
    nodes near coords style={font=\normalsize,yshift=-15pt}
    ]
    coordinates {
    (0,20.2)[20.2](1,21.5)[21.5](2,21.7)[21.7](3,20.6)[20.6](4,20.6)[20.6](5,19.1)[19.1](6,19.4)[19.4](7,18.7)[18.7]
    };
    \addlegendentry{Chinese-to-English}

\addplot[
    color=blue,
    mark=square,
    nodes near coords style={font=\normalsize},
    ]
    coordinates {
    (0,9.5)[9.5](1,20.6)[20.6](2,20.1)[20.1](3,20.3)[20.3](4,21.1)[21.1](5,22.3)[22.3](6,20.3)[20.3](7,22.5)[22.5]
    };
    \addlegendentry{English-to-Chinese}

\end{axis}
\end{tikzpicture}

\end{subfigure}

\vspace{0.5cm}

\begin{subfigure}[b]{0.49\textwidth}
\centering
\begin{tikzpicture}
\begin{axis}[
    legend columns=1,legend style={font=\small},
    title={\bf English$\leftrightarrow$Ukrainian},
    xlabel={\bf Sizes of Parallel Data},
    ylabel={\bf BLEU},
    xmin=-0.2, xmax=7.2,
    ymin=2, ymax=50,
    ytick={5,10,15,20,25,30,35,40,45},
    xticklabels={0, 10k,50k,100k,500k,1M,5M,10M},
    xtick={0,1,2,3,4,5,6,7},
    legend pos=south east,
    ymajorgrids=true,
    grid style=dashed,
    height=2.3in,
    width=1.0\textwidth,
    nodes near coords,
    point meta=explicit symbolic
]

\addplot[
    color=red,
    mark=triangle,
    mark options={scale=1.5},
    nodes near coords style={font=\normalsize},
    ]
    coordinates {
    (0,36.6)[36.6](1,36.3)[36.3](2,36.1)[36.1](3,35.6)[35.6](4,35.0)[35.0](5,35.9)[35.9](6,35.8)[35.8](7,37.2)[37.2]
    };
    \addlegendentry{Ukrainian-to-English}

\addplot[
    color=blue,
    mark=square,
    nodes near coords style={font=\normalsize},
    ]
    coordinates {
    (0,18.1)[18.1](1,14.9)[14.9](2,16.4)[16.4](3,16.0)[16.0](4,25.2)[25.2](5,24.0)[24.0](6,24.3)[24.3](7,23.9)[23.9]
    };
    \addlegendentry{English-to-Ukrainian}

\end{axis}
\end{tikzpicture}
\end{subfigure}
\begin{subfigure}[b]{0.49\textwidth}
\centering
\begin{tikzpicture}
\begin{axis}[
    legend columns=1,legend style={font=\small},
    title={\bf English$\leftrightarrow$Hebrew},
    xlabel={\bf Sizes of Parallel Data},
    ylabel={\bf BLEU},
    xmin=-0.2, xmax=7.2,
    ymin=0, ymax=40,
    ytick={5,10,15,20,25,30,35,40,45},
    xticklabels={0, 10k,50k,100k,500k,1M,5M,10M},
    xtick={0,1,2,3,4,5,6,7},
    legend pos=south east,
    ymajorgrids=true,
    grid style=dashed,
    height=2.3in,
    width=1.0\textwidth,
    nodes near coords,
    point meta=explicit symbolic
]

\addplot[
    color=red,
    mark=triangle,
    mark options={scale=1.5},
    nodes near coords style={font=\normalsize},
    ]
    coordinates {
    (0,17.2)[17.2](1,19.7)[19.7](2,23.8)[23.8](3,25.8)[25.8](4,30.0)[30.0](5,31.8)[31.8](6,34.1)[24.1](7,35.0)[35.0]
    };
    \addlegendentry{Hebrew-to-English}

\addplot[
    color=blue,
    mark=square,
    nodes near coords style={font=\normalsize},
    ]
    coordinates {
    (0,1.5)[1.5](1,4.6)[4.6](2,8.3)[8.3](3,9.0)[9.0](4,13.6)[13.6](5,15.6)[15.6](6,19.8)[19.8](7,8.4)[8.4]
    };
    \addlegendentry{English-to-Hebrew}

\end{axis}
\end{tikzpicture}
\end{subfigure}

\caption{BLEU scores of bi-directional translation results using Llama2-7b based translation models across multiple language pairs, where English, German, Chinese, Ukrainian, and Hebrew are sorted based on the level of available resources, with English having the highest resources and Hebrew the least in the pertaining stage.}
\label{fig:otherlanguages} 

\end{figure*}
\begin{table*}[h!]

\resizebox{\textwidth}{!}{
\begin{tabular}{lcccccccc}
\toprule
\multicolumn{1}{l}{\multirow{2}{*}{\bf Model $\downarrow$}} & \multicolumn{2}{c}{\bf English-to-German} & \multicolumn{2}{c}{\bf English-to-Chinese} & \multicolumn{2}{c}{\bf English-to-Ukrainian} & \multicolumn{2}{c}{\bf English-to-Hebrew} \\ \cmidrule{2-3} \cmidrule{4-5} \cmidrule{6-7} \cmidrule{8-9}
\multicolumn{1}{l}{}                       & \bf BLEU            & \bf COMET-DA            & \bf BLEU             & \bf COMET-DA            & \bf BLEU              &  \bf COMET-DA             &  \bf BLEU            &  \bf COMET-DA            \\ \midrule
LLM-SFT                                 & 28.3            & 84.5        & 22.5      & 78.4   & \bf 25.2  &  \bf 88.8     & \bf 19.8   & \bf 78.6                \\  \cdashline{1-9}\noalign{\vskip 0.5ex}
ALMA-7B  & \textbf{30.2}     & 85.3       & 29.5     & 83.0  & 10.6    & 88.3     & 0.7      & 57.4     \\
TowerInstruct-7B      & 29.8       &  \bf 85.7        & \bf 31.4    & \bf 83.5                & 12.4      & 87.0                   & 2.5             & 44       \\
\bottomrule
\end{tabular}}

\caption{Translation Performance on four English-to-X directions of existing multilingual large language models. LLM-SFT presents the best result in Figure~\ref{fig:otherlanguages}. The results suggest that existing MLLMs significantly improve the translation capabilities for pretrained non-English languages. However, they have not yet focused on enhancing the translation performance for low-resource languages.}
\label{tab:mllms_results}

\end{table*}

\section{New Challenges}


Within the research field of LLMs, two pressing challenges arise. 
The first challenge concerns the translation quality for language pairs that are underrepresented during the pretraining stage, possessing significantly less data volume compared to major languages. In our case, approximately $90\%$ of the data in Llama2-7b is English \cite{touvron2023llama2}, leading to a highly skewed data distribution. The second challenge involves the evaluation of translation quality. Automatic evaluation metrics, such as BLEU or COMET, may not fully correspond with human evaluation standards \cite{liu-etal-2023-g,mao2023gpteval}, complicating the accurate assessment of LLMs.

\subsection{Pretraining Resource Imbalance}
\label{sec:mllms}

We conduct an experiment using Llama-7b-based translation models on four translation pairs from the WMT23 tasks: German-to-English, Chinese-to-English, Ukrainian-to-English, and Hebrew-to-English translations. According to the technical report of Llama2 \cite{touvron2023llama2}, the distribution of the five languages in the dataset of Llama2 is as follows: $89.70\%$ for English, $0.17\%$ for German, $0.13\%$ for Chinese, $0.07\%$ for Ukrainian, and less than $0.005\%$ for Hebrew. The results are presented in Figure~\ref{fig:otherlanguages}, and we observe:

\begin{itemize}[leftmargin=*,topsep=0.1em,itemsep=0.1em,parsep=0.1em]
    \item \textbf{Translation performance is significantly impacted by the available resources for each language.}  
    The X-to-English tasks, with the highest resource of English, consistently achieve stable translation performance compared to other directions, as evidenced by a smooth performance curve (except for Hebrew-to-English) with varying amounts of parallel data. Conversely, Hebrew, with the least resource in Llama2 pretraining data, exhibits improved translation performance as parallel data increases. This observation underscores the importance of diverse and balanced datasets for pretraining LLMs to ensure equitable performance across languages.

    \item \textbf{For low-resource pretraining data languages, a substantial change in slope is observed as the amount of parallel data increases.}
    The English-to-X tasks, except English-to-Hebrew, exhibit fluctuating performance curves as the volume of parallel data increases. Specifically, the curve for English-to-German experiences a sharp rise from $0$k to $10$k parallel data, a pattern also observed in English-to-Chinese. Both German and Chinese account for more than $0.1\%$ of the pretraining data distribution. The English-to-Ukrainian curve displays a notable inflection point between $100$k and $500$k, with Ukrainian constituting $0.07\%$ of the data distribution. In contrast, the English-to-Hebrew curve remains smooth, as Hebrew does not fall within the languages that comprise more than $0.005\%$ of the data distribution. These points of significant increase may serve as indicators of the emergent translation capabilities of LLMs.

\end{itemize}


The above findings suggest that addressing the challenges of low-resource pretrained languages and identifying inflection points where LLMs' capabilities emerge will be essential to enhance the effectiveness of various translation directions. 


\subsubsection{Performance of Existing MLLMs}



Numerous studies have recently focused on adapting English-centric Large Language Models (LLMs) into Multilingual LLMs (MLLMs) \cite{xu2024survey, qin2024multilingual}. We employ two publicly accessible MLLMs, ALMA-7B \cite{xu2024a} and TowerInstruct-7B \cite{tower_llm_2024}, and directly evaluate them on the four English-to-X translation routes depicted in Figure~\ref{fig:otherlanguages}. Both models consistently utilize monolingual data from various languages for continued pretraining, followed by fine-tuning with high-quality parallel data, based on the Llama2-7B model. Consequently, these two models successfully enhance the translation quality of pretrained languages such as German and Chinese.

\textbf{MLLMs exhibit a degree of translation generalization.}
Despite not being pretrained on Ukrainian, ALMA-7B and TowerInstruct-7B achieve high COMET-DA scores of 88.3 and 87.0, respectively, indicating successful English-to-Ukrainian translations with high semantic consistency. 
We posit that this success is attributable to their training on Russian texts, a linguistically similar language to Ukrainian. 
Nevertheless, both models demonstrate limited translation capabilities for Hebrew due to insufficient training data in this language. 
These findings underscore the necessity for further research and development to enhance the LLMs for low-resource languages.


\subsection{Evaluation Issues}
\label{sec:evaluationissues}

We conducted an assessment of three representative systems using the WMT2023 Discourse-Level Literary Translation Chinese-English Testset.\footnote{\url{https://www2.statmt.org/wmt23/literary-translation-task.html}} 
These include: the \textit{Commercial Translation System}, Google Translate,\footnote{\url{https://translate.google.com}} known for its superior translation performance; the \textit{Commercial LLM Systems}, specifically the GPT-4 (8K) API,\footnote{\url{https://platform.openai.com}.} recognized for its comprehensive context modeling capabilities \cite{ouyangtraining-instructGPT,wang2023document}; and the \textit{Open-sourced LLM Models}, particularly Llama2-7b (4K) \cite{touvron2023llama2}, optimized for document-level translation using a 200K general-domain document-level training set. For evaluation, we utilized both automatic and human methods. 
The former uses document-level sacreBLEU (d-BLEU) \cite{liu-etal-2020-multilingual-denoising}, while the latter adopts multidimensional quality metrics (MQM) \cite{lommel2014multidimensional} to fit the literary translation context. 
Table~\ref{tab:auto-vs-human} presents a comparative performance analysis of these systems.

\begin{itemize}[leftmargin=*,topsep=0.1em,itemsep=0.1em,parsep=0.1em]
\item \textbf{A moderate negative correlation exists between human and automatic evaluations.}
A Pearson correlation coefficient of -0.53 indicates a divergence between the two evaluation methods. This discrepancy suggests that human and automatic evaluations can provide complementary insights into document-level translation quality. The findings underscore the importance of combining both evaluation methods and highlight the potential limitations of current automatic evaluation approaches in assessing machine translation systems.

\item \textbf{The need for human-aligned evaluation in LLMs.}
The application of LLMs for translation tasks emphasizes the need for evaluation methods that accurately reflect human-like or human-preferred translation quality. The observed divergence between human and automatic evaluations in our study suggests that current automatic metrics may not fully capture the nuances appreciated by human evaluators. This calls for further research to develop and refine evaluation methods better aligned with human preferences and expectations in translation, especially as we continue to enhance the complexity and capabilities of language models. 

\end{itemize}

The above findings highlight the importance of advancing evaluation methodologies that align with human preferences for LLM translation.

\begin{table}[t]
    \centering
    \begin{tabular}{c r r}
     \toprule
     {\bf System} & {\color{red} \bf MQM} & {\color{blue}\bf d-BLEU}\\
    \midrule
   GPT-4 & \cellcolor{red!30} 54.81 & \cellcolor{blue!15} 43.7\\
   Llama2-MT & \cellcolor{red!15} 28.40  & \cellcolor{blue!5} 43.1\\
   Google & \cellcolor{red!5} 22.66  & \cellcolor{blue!30} 47.3\\
    \bottomrule
    \end{tabular}
    \caption{The comparison between human (MQM) vs. automatic (d-BLEU) evaluation methods over three representative systems on the Chinese-to-English translation task, with a color scale to denote the ranking. The Pearson correlation coefficient between MQM and d-BLEU is $-0.53$.}
    \label{tab:auto-vs-human}
\end{table}

\section{Conclusions}

Our research highlights several key findings. 
On the positive side, LLMs have effectively removed the dependence on bilingual data for translation into major languages.
Additionally, despite being fine-tuned solely on sentence-level translation pairs, LLMs demonstrate impressive capabilities in handling long sentences and even document-level translations. 
However, challenges persist. 
LLMs face difficulties in adapting to multi-domain tasks and predicting rare words. 
Our experiments suggest that larger models possess greater potential to acquire multi-domain translation knowledge, offering a promising path to mitigating the domain shift issue. 
Yet, performance for low-resource languages remains suboptimal, and inference delays pose a significant bottleneck. 
Addressing these limitations, particularly improving translation for underrepresented languages, will be crucial. 
Our findings prove that leveraging bilingual corpora more efficiently could offer a viable solution. 
Furthermore, the need for more robust, human-aligned evaluation metrics remains an urgent area for future research.





\section*{Acknowledgments}

This work was supported in part by the Science and Technology Development Fund, Macau SAR (Grant No. FDCT/0070/2022/AMJ, the mainland China collaboration project, China Strategic Scientific and Technological Innovation Cooperation Project Grant No. 2022YFE0204900), the Science and Technology Development Fund, Macau SAR (Grant No. FDCT/060/2022/AFJ, the mainland China collaboration project, National Natural Science Foundation of China Grant No. 62261160648), the Multi-year Research Grant from the University of Macau (Grant No. MYRG-GRG2024-00165-FST), and the Tencent AI Lab Rhino-Bird Gift Fund (Grant No. EF2023-00151-FST).

\bibliography{anthology,custom}
\bibliographystyle{acl_natbib}

\clearpage

\appendix

\section{Detailed Experimental Settings}

\subsection{Models}



In this section, we provide a detailed description of the models and settings used in our experiments.

\begin{itemize}[leftmargin=*,topsep=0.1em,itemsep=0.1em,parsep=0.1em]
    \item For the LLM-SFT models, we employ Llama2-7B and Llama2-13B as the backbone models. During the training process, a learning rate of 3e-4 is used for continued pre-training (CT), while a learning rate of 2e-5 is applied for supervised fine-tuning (SFT) \cite{touvron2023llama2}.
    \item For the Enc2Dec models, we utilize the transformer-base architecture for translation tasks with 500k or fewer parallel texts, and the transformer-big architecture for tasks with more than 500k parallel texts \cite{pang-etal-2024-monmt-modularly}. During the training phase, a learning rate of 1e-5 and a batch size of 8,192 maximum tokens are used. We evaluate the model every 1,000 update steps and terminate the training process when the validation performance plateaus or worsens, with a patience of 20 \cite{pang2023rethinking}. The Adam optimizer is employed to optimize model parameters, with $\beta_{1}=0.9$, $\beta_{2}=0.98$, and $\epsilon=10^{-9}$ \cite{kingma2014adam}.
    \item In Section~\ref{sec:evaluationissues}, Llama2-MT is trained on an open-source, document-level training set \cite{wang-etal-2023-findings}.\footnote{\url{https://www2.statmt.org/wmt23/literary-translation-task.html}}
    \item ALMA-7B and TowerInstruct-7B are obtained from the open-source HuggingFace community.\footnote{\url{https://huggingface.co/haoranxu/ALMA-7B}, \url{https://huggingface.co/Unbabel/TowerInstruct-7B-v0.2}}
    Default settings are used to generate translation hypotheses \cite{xu2024a,tower_llm_2024}.
\end{itemize}

In this section, we have provided a comprehensive overview of the models and settings employed in our study, ensuring that our methodology is clear and reproducible.

\subsection{Definitions of LLM training}

In this section, we provide clear definitions of LLM training methods, including instruction tuning, supervised fine-tuning (SFT), and continued pre-training (CPT).

\begin{itemize}[leftmargin=*,topsep=0.1em,itemsep=0.1em,parsep=0.1em]

\item  \textbf{Instruction Tuning}: Instruction tuning is a process of fine-tuning large language models to follow specific instructions provided in the input \cite{openai2023gpt4,alpaca}. This is done by training the model on a dataset containing examples with both instructions and corresponding correct responses. The goal is to make the model more controllable and useful by enabling it to understand and respond to explicit instructions given by the user.

\item  \textbf{Supervised Fine-Tuning}: Supervised fine-tuning is the process of adapting a pre-trained language model to a specific task or domain using labeled data. This involves training the model on a dataset containing input-output pairs, where the outputs are the ground truth or correct responses \cite{jiao2023parrot}. The model learns to generate appropriate responses based on the input by minimizing the difference between its predictions and the ground truth. Supervised fine-tuning helps improve the model's performance on tasks like text classification, sentiment analysis, and question-answering \cite{fang2023chatgpt,pang2024anchor}.

\item  \textbf{Continued Pre-training}: Continuous pre-training refers to the ongoing process of training a language model on a large corpus of text, often from diverse sources, to learn general language understanding \cite{cossu2022continual,gupta2023continual}. This is done before any fine-tuning or task-specific training. The idea is to keep updating the model's knowledge and understanding of language as new data becomes available, making it more robust and adaptable. 

\end{itemize}

\subsection{Instruction Formats}
\begin{table}[h]
    \centering
    \begin{tabular}{cp{6cm}} 
    \toprule
        \bf Type & \multicolumn{1}{l}{\bf Data Format} \\ \midrule
          \bf SFT   & Below is an instruction that describes a task, paired with an input that provides further context. Write a response that appropriately completes the request. \\ \cdashline{2-2}\noalign{\vskip 0.5ex}
             & \#\#\# Instruction: Translate the following sentences from German to English. \\ \cdashline{2-2}\noalign{\vskip 0.5ex}
             & \#\#\# Input: Haben Sie einen Blick auf andere Restaurants in der Nähe mit ähnlicher Küche geworfen? \\ \cdashline{2-2}\noalign{\vskip 0.5ex}
             & \#\#\# Response: Have you looked at other nearby restaurants with similar cuisine? \\ \midrule

        \bf CPT & [German]: Haben Sie einen Blick auf andere Restaurants in der Nähe mit ähnlicher Küche geworfen? [English]: Have you looked at other nearby restaurants with similar cuisine? \\ \bottomrule
            
    \end{tabular}
    \caption{Data formats utilized for training LLMs, where SFT represents supervised finetuning, and CPT denotes continued pertaining.}
    \label{tab:insdata}
\end{table}


Table~\ref{tab:insdata} showcases two unique data formats that are employed for the purpose of constructing bilingual pairs, which are essential for training Large Language Models (LLMs). The first format adheres to the design principles of the Alpaca dataset, as described in the literature \cite{alpaca}. This format has been widely used and accepted in various research studies and applications. On the other hand, the second format involves concatenating two pairs of text, each representing a different language. To distinguish between the languages in this concatenated format, language tags are incorporated \cite{zhu2023multilingual}.

\subsection{Rare Word Precision}

\begin{algorithm}[h]
    \caption{Word Precision and Delete Rates}\label{alg:cap}
    \small
    \begin{algorithmic}[1]
        \Require The source texts $T_{src}$, the target texts $T_{tgt}$, and the source-to-target alignments of hypothesis pairs $\mathcal{A}_{hyp}$, with identical sorted index; The source word frequency $word2freq$;
        \Ensure Word precision $\mathcal{P}$ and deletion rates $\mathcal{D}$;
        \State $\mathcal{P}, \mathcal{D} \leftarrow$ Initialized as an zero list;
        \ForEach {$t_{src}$, $t_{tgt}$, ${a}_{hyp}$ in ($T_{src}$, $T_{tgt}$, $\mathcal{A}_{hyp}$)}
            \ForEach{source word $w_{s}$ in $t_{src}$}
                \If {$w_{s}$ not in $a_{hyp}$}
                    \State $\mathcal{D}_{w_{s}} +=$ 1;
                    \State continue;
                \EndIf
                \State List of target words $W^{t}_{hyp} =$ $a_{hyp}$($w_{s}$);
                \State $c^{t}_{hyp} \leftarrow$ Length of $W^{t}_{hyp}$;
                \State $c^{t}_{ref} \leftarrow$ $0$;
                \ForEach{$rw$ in $W^{t}_{hyp}$}
                    \ForEach{$tw$ in $t_{tgt}$}
                        \If {$tw$ == $rw$}
                            \State $c^{t}_{ref} +=$ 1;
                        \EndIf
                    \EndFor
                \EndFor

                \State $minc =$ Min($c^{t}_{ref}$,$c^{t}_{hyp}$);
                \State $maxc =$ Max($c^{t}_{ref}$,$c^{t}_{hyp}$);
                \State $\mathcal{P}_{w_{s}} +=$ Min($\frac{minc}{maxc}$,$1$);
                
            \EndFor
        \EndFor
       \State $\mathcal{P}_{w} = \mathcal{P}_{w}/word2freq(w)$;
       \State $\mathcal{D}_{w} = \mathcal{D}_{w}/word2freq(w)$;
       \State \Return {$\mathcal{P}, \mathcal{D}$}.
    \end{algorithmic}
\end{algorithm}

First, we calculate the source word type frequency using the entire WMT23 bilingual corpus, which comprises $295,805,439$ translation pairs. Subsequently, we concatenate the translation results with the corpus and employ \emph{FastAlign}\footnote{\url{https://github.com/clab/fast_align}} to determine the alignment information. In this manner, we adhere to the methodologies employed in previous studies \cite{koehn-haddow-2012-interpolated,koehn-knowles-2017-six} to compute the word precision and deletion ratio for each source word, as demonstrated in Algorithm~\ref{alg:cap}.

\subsection{Human Evaluation}

This section provide the detail information about the human evaluation in Section~\ref{sec:evaluationissues}.
Following the industry-endorsed criteria of \newcite{wang-etal-2023-findings}, the human evaluation was performed by professional translators using an adaptation of the multidimensional quality metrics (MQM) framework \cite{lommel2014multidimensional}.

\section{Further Experimental Resutls}
\label{sec:appendix}






\subsection{Alignment}

\begin{table*}[h]
\centering
\scalebox{0.77}{

\begin{tabular}{l rrrrrrrrrrrrrrrr}
\toprule
\bf Layer  & \bf 1    &  \bf 2     & \bf 3     & \bf 4    & \bf 5   & \bf 6    & \bf 7    & \bf 8     & \bf 9     &  \bf 10    & \bf 11    & \bf 12  &  \bf 13    & \bf 14   &  \bf 15   & \bf 16   \\ \midrule
T      & $\_$in  & .     & ute   & .    & .   & $\_$in  & $\_$den & $\_$ich  & $\_$in   & $\_$in   & .     & .   & $\_$den  & $\_$den & $\_$den & $\_$den \\
od     & he   & $\_$Park & ute   & $\_$den & ute & $\_$in  & $\_$den & $\_$den  & $\_$in   & $\_$in   & .     & .   & he    & $\_$den & $\_$den & $\_$den \\
ay     & he   & .     & ute   & $\_$den & ute & $\_$in  & $\_$den & $\_$Park & $\_$in   & $\_$in   & .     & ute & he    & $\_$den & $\_$den & $\_$den \\
,      & .    & .     & ute   & $\_$den & ute & $\_$den & $\_$den & $\_$Park & He    & $\_$in   & .     & ute & $\_$in   & $\_$den & $\_$den & $\_$den \\
$\_$I     & $\_$in  & $\_$den  & .     & $\_$den & ute & $\_$in  & $\_$den & He    & $\_$in   & $\_$in   & .     & ute & $\_$den  & .    & $\_$den & $\_$den \\
'      & $\_$in  & He    & ute   & $\_$den & ute & He   & $\_$den & He    & He    & $\_$in   & He    & ute & .     & .    & $\_$den & $\_$den \\
m      & $\_$in  & $\_$Park & ute   & $\_$den & he  & $\_$in  & $\_$den & $\_$ge   & $\_$in   & $\_$in   & $\_$in   & ute & he    & .    & $\_$den & $\_$den \\
$\_$going & $\_$in  & .     & .     & $\_$den & he  & $\_$in  & $\_$den & $\_$ge   & $\_$in   & $\_$in   & $\_$in   & ute & $\_$Park & .    & .    & $\_$den \\
$\_$to    & $\_$in  & $\_$den  & ute   & $\_$den & he  & $\_$in  & $\_$den & $\_$ich  & $\_$in   & $\_$in   & .     & ute & $\_$den  & .    & .    & $\_$den \\
$\_$the   & $\_$in  & $\_$in   & ute   & $\_$den & .   & $\_$den & $\_$den & $\_$den  & $\_$in   & $\_$in   & .     & ute & $\_$den  & .    & .    & $\_$den \\
$\_$park  & $\_$in  & .     & .     & $\_$den & .   & $\_$in  & $\_$den & $\_$Park & $\_$in   & $\_$in   & $\_$in   & ute & $\_$Park & .    & .    & $\_$den \\
.      & $\_$in  & $\_$in   & .     & He   & .   & .    & $\_$den & $\_$Park & He    & .     & He    & ute & .     & .    & .    & $\_$den \\ \midrule
\bf Layer  & \bf 17   & \bf 18    & \bf 19    & \bf 20   & \bf 21  & \bf 22   & \bf 23   & \bf 24    & \bf 25    & \bf 26    & \bf 27    & \bf 28  & \bf 29    & \bf 30   & \bf 31   & \bf 32   \\ \midrule 
T      & $\_$in  & $\_$in   & $\_$Park & .    & $\_$in & $\_$den & $\_$in  & .     & $\_$in   & .     & $\_$in   & .   & $\_$in   & $\_$ich & $\_$den & $\_$in  \\
od     & $\_$den & $\_$in   & $\_$in   & .    & $\_$in & $\_$den & $\_$in  & .     & $\_$in   & .     & $\_$in   & .   & $\_$in   & $\_$ich & .    & $\_$in  \\
ay     & $\_$den & $\_$in   & $\_$in   & $\_$den & $\_$in & $\_$den & $\_$in  & ute   & .     & .     & $\_$in   & .   & $\_$in   & $\_$ich & .    & $\_$in  \\
,      & $\_$in  & $\_$in   & $\_$in   & $\_$den & $\_$in & $\_$den & $\_$in  & .     & .     & .     & $\_$in   & .   & $\_$in   & .    & He   & $\_$in  \\
$\_$I     & $\_$ich & $\_$in   & $\_$in   & $\_$den & $\_$in & $\_$den & $\_$in  & .     & .     & .     & $\_$in   & .   & $\_$in   & $\_$ich & $\_$in  & $\_$in  \\
'      & $\_$ich & $\_$in   & $\_$ich  & .    & $\_$in & $\_$den & $\_$in  & He    & ute   & $\_$Park & $\_$in   & .   & $\_$in   & $\_$ich & $\_$den & $\_$in  \\
m      & $\_$den & $\_$in   & $\_$den  & .    & $\_$in & $\_$den & $\_$in  & .     & he    & .     & $\_$in   & .   & $\_$in   & $\_$ich & $\_$den & $\_$in  \\
$\_$going & $\_$ich & $\_$in   & $\_$den  & .    & $\_$in & $\_$den & $\_$in  & .     & he    & .     & $\_$in   & .   & $\_$in   & $\_$ich & $\_$den & $\_$in  \\
$\_$to    & $\_$in  & $\_$in   & $\_$in   & .    & $\_$in & $\_$den & $\_$in  & .     & he    & .     & $\_$in   & .   & $\_$in   & $\_$ich & $\_$den & $\_$in  \\
$\_$the   & $\_$in  & $\_$in   & $\_$in   & .    & $\_$in & $\_$den & $\_$in  & .     & $\_$in   & .     & $\_$in   & .   & $\_$in   & .    & He   & $\_$in  \\
$\_$park  & $\_$ich & $\_$in   & $\_$den  & $\_$den & $\_$in & $\_$den & $\_$in  & .     & $\_$Park & .     & $\_$in   & $\_$in & $\_$in   & $\_$ich & He   & $\_$in  \\
.      & $\_$in  & $\_$Park & .     & $\_$den & $\_$in & $\_$den & .    & .     & .     & .     & $\_$Park & $\_$in & .     & $\_$den & He   & $\_$in \\
\bottomrule
\end{tabular}
}
\caption{Word alignment induced from target-to-source attention weights for each layer in LLM-SFT German-to-English translation model. The translation model is supervised-finetuned on $100$k parallel data. The left row is a target English sentence ``Today, I'm going to the park.'', and its source German sentence is ``Heute gehe ich in den Park.''. Both sentences are tokenized by the Llama2 tokenizer. We observe that target tokens tend to attend the same source tokens within each layer.}
\label{tab:alignmentexample}
\end{table*}

Table~\ref{tab:alignmentexample} showcases the word alignment outcomes derived from each layer of the LLM-SFT translation model, which was trained on $100$k German-to-English datasets. 
The findings indicate a tendency for target tokens to align with the same source tokens throughout all layers.

\subsection{Beam Search}

\pgfplotsset{superb legend/.style={legend style                                = {draw=none,
                 legend columns                          = 3,
                 /tikz/every even column/.append style   = {column sep=0.5cm,
                 text width=7em},
                 /tikz/every odd column/.append style    = {column sep=0.15cm,
                  text width=7em},
                 }}}

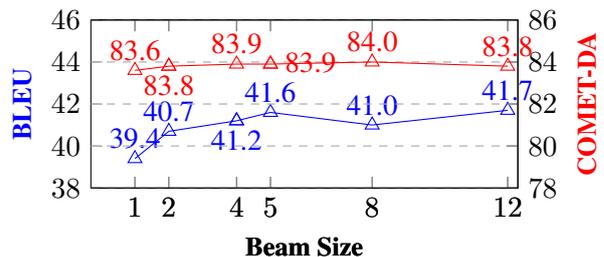
\begin{figure}[ht]

    \centering

\begin{tikzpicture}
\pgfplotsset{
every axis legend/.append style={at={(0.5,1.03)},anchor=south},
}

\begin{axis}[
    legend columns=3,legend style={font=\small}, label style={font=\small},
    xlabel={\bf Beam Size},
    ylabel={\bf {\color{blue}BLEU}},
    xmin=-0.3, xmax=12.3,
    ymin=38, ymax=46,
    xtick={1,2,4,5,8,12},
    ymajorgrids=true,
    grid style=dashed,
    height=1.5in,
    width=0.45\textwidth,
    nodes near coords
]

\addplot[
    color=blue,
    mark=triangle,
    mark options={scale=1.5},
    point meta=explicit symbolic,
    ]
    coordinates {
    (1,39.4)[39.4](2,40.7)[40.7](4,41.2)[](5,41.6)[41.6](8,41.0)[41.0](12,41.7)[41.7]
    };

\addplot[
    color=blue,
    mark=triangle,
    nodes near coords style={yshift=-15pt},
    mark options={scale=1.5},
    point meta=explicit symbolic,
    forget plot
    ]
    coordinates {
    (4,41.2)[41.2]
    };

\end{axis}

\begin{axis}[
    legend columns=3,legend style={font=\small}, label style={font=\small},
    axis y line*=right,
    xlabel={\bf Beam Size},
    ylabel={\bf {\color{red}COMET-DA}},
    xmin=-0.3, xmax=12.3,
    ymin=78, ymax=86,
    xtick={1,2,4,5,8,12},
    ymajorgrids=true,
    grid style=dashed,
    height=1.5in,
    width=0.45\textwidth,
    nodes near coords
]

\addplot[
    color=red,
    mark=triangle,
    mark options={scale=1.5},
    point meta=explicit symbolic,
    ]
    coordinates {
    (1,83.6)[83.6](2,83.8)[](4,83.9)[83.9](5,83.9)[](8,84.0)[84.0](12,83.8)[83.8]
    };

\addplot[
    color=red,
    mark=triangle,
    mark options={scale=1.5},
    nodes near coords style={yshift=-15pt},
    point meta=explicit symbolic,
    forget plot
    ]
    coordinates {
    (2,83.8)[83.8]
    };

\addplot[
    color=red,
    mark=triangle,
    mark options={scale=1.5},
    nodes near coords style={xshift=15pt},
    nodes near coords style={yshift=-7pt},
    point meta=explicit symbolic,
    forget plot
    ]
    coordinates {
    (5,83.9)[83.9]
    };


\end{axis}

\end{tikzpicture}

\caption{BLEU and COMET-DA scores with beam size for LLM-SFT-100k in Section~\ref{sec:c2}.}
\label{fig:app-bs}

\end{figure}


Figure~\ref{fig:app-bs} shows the German-to-English translation performance on generaltest2023 of WMT23 with increasing beam size.
The results show that increasing the beam size enhances surface-level similarity to the ground truth, but has minimal impact on semantic-level similarity.




\end{document}